\documentclass[10pt,twocolumn,letterpaper]{article}
\usepackage{smg}
\usepackage{times}
\usepackage[dvipsnames]{xcolor}
\usepackage{adjustbox} %
\usepackage{amsfonts}
\usepackage{amsmath}
\usepackage{amssymb}
\usepackage{array}
\usepackage{arydshln}
\usepackage{bbm}
\usepackage[numbers,sort&compress]{natbib}
\usepackage{booktabs}
\usepackage{comment}
\usepackage{dblfloatfix}
\usepackage{dsfont}
\usepackage{float}
\usepackage{graphicx,wrapfig,lipsum}
\usepackage{mathtools,nccmath}
\usepackage{multirow}
\usepackage{pifont}
\usepackage{relsize}
\usepackage{url}
\usepackage{subcaption}
\usepackage[accsupp]{axessibility}

\usepackage[pagebackref=true,breaklinks=true,colorlinks,bookmarks=false]{hyperref}
\usepackage[capitalize]{cleveref}

\makeatletter
\renewcommand{\paragraph}{%
  \@startsection{paragraph}{4}%
  {\z@}{0.25em}{-1em}%
  {\normalfont\normalsize\bfseries}%
}
\makeatother

\newcommand{\cmark}{\ding{51}}%
\newcommand{\xmark}{\ding{55}}%

\makeatletter
\DeclareRobustCommand\onedot{\futurelet\@let@token\@onedot}
\def\@onedot{\ifx\@let@token.\else.\null\fi\xspace}
\def\eg{\emph{e.g}\onedot}

\newlength{\fHeight}
\newcommand{\faBird}{\includegraphics[height=\fHeight]{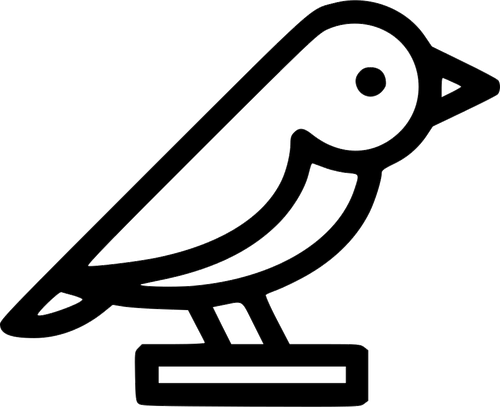}}

\newcommand{\faCat}{\includegraphics[height=\fHeight]{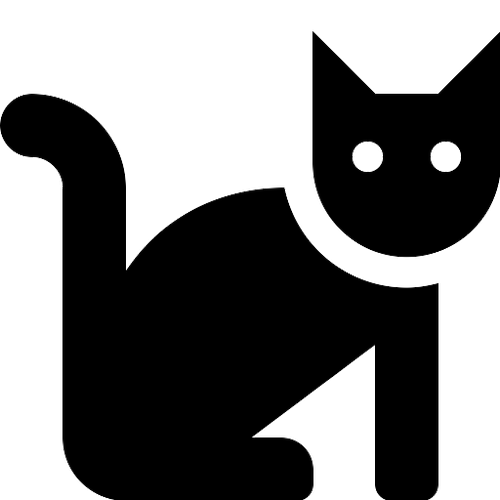}}
\newcommand{\faCow}{\includegraphics[height=\fHeight]{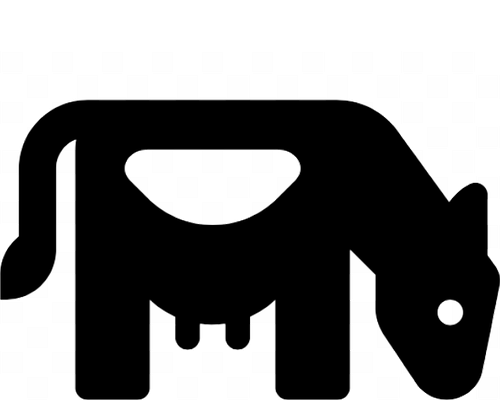}}

\newcommand{\faDog}{\includegraphics[height=\fHeight]{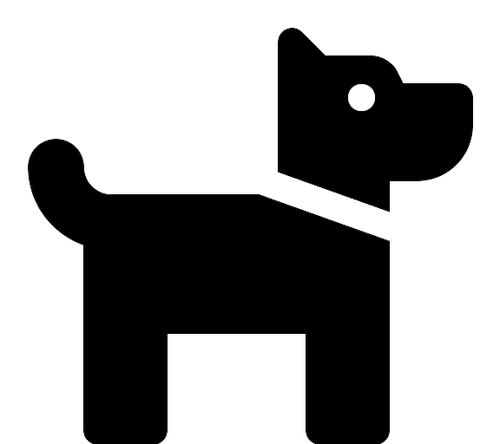}}

\newcommand{\faHorse}{\includegraphics[height=\fHeight]{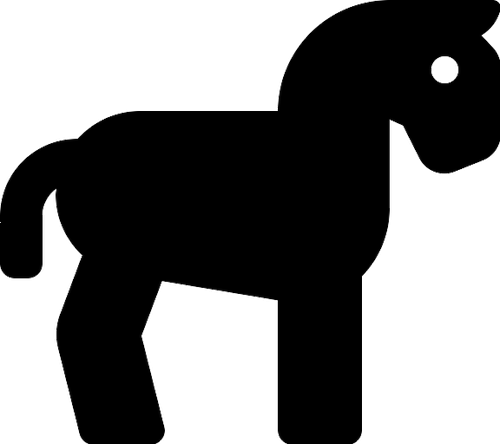}}

\newcommand{\faSheep}{\includegraphics[height=\fHeight]{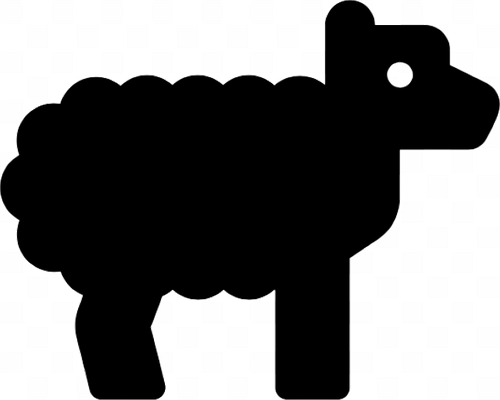}}

\newcommand{\faCircle}{\includegraphics[height=2ex]{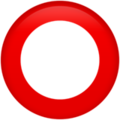}}

\smgfinalcopy
\def\smgPaperID{10759}
\ifsmgfinal\fi

\title{What does CLIP know about a red circle? \\
Visual pr\faCircle mpt engineering for VLMs}

\author{
Aleksandar Shtedritski
\quad
Christian Rupprecht
\quad
Andrea Vedaldi
\\[0.3em]
Visual Geometry Group, University of Oxford\\
{\tt\small \{suny, chrisr, vedaldi\}@robots.ox.ac.uk}
}

\begin{document}
\maketitle
\ifsmgfinal\thispagestyle{empty}\fi

\begin{abstract}
Large-scale Vision-Language Models, such as CLIP, learn powerful image-text representations that have found numerous applications, from zero-shot classification to text-to-image generation.
Despite that, their capabilities for solving novel discriminative tasks via prompting fall behind those of large language models, such as GPT-3.
Here we explore the idea of visual prompt engineering for solving computer vision tasks beyond classification by editing in image space instead of text.
In particular, we discover an emergent ability of CLIP, where, by simply drawing a red circle around an object, we can direct the model's attention to that region, while also maintaining global information.
We show the power of this simple approach by achieving state-of-the-art in zero-shot referring expressions comprehension and strong performance in keypoint localization tasks.
Finally, we draw attention to some potential ethical concerns of large language-vision models.
\end{abstract}

\section{Introduction}%
\label{sec:intro}

Large Language Models  (LLMs) such as GPT-2/3~\cite{brown2020language_gpt3, radford2019language-gpt2} and ChatGPT~\cite{chatgpt} have demonstrated surprising emerging behaviours.
For example, these models can perform language translation without being explicitly trained for it, in a zero-shot manner.
This can be partially explained by the fact that occurrences of the desired behaviours, such as translating between two languages, naturally occur in their enormous training corpus, which is, essentially, the Internet.

Interesting emergent behaviours have been observed in large Vision-Language Models (VLMs) like CLIP~\cite{radford2021learning} too.
For example, CLIP can be used for zero-shot classification by checking the compatibility of a given image with prompts such as ``an image of a $X$'', where $X$ is one of a set of class hypotheses to be tested.

Emergent behaviours are elicited by supplying suitably crafted inputs to the VLMs, often called \emph{prompts}.
As in the example above, researchers have mostly focused on \emph{engineering textual prompts}, manipulating the textual input of the model.
This approach is inspired by LLMs, where manipulating the textual modality is the only available option.
However, VLMs are inherently multimodal and offer the possibility of manipulating both modalities, textual and visual.
While the textual modality is the natural choice for expressing semantics, the visual modality can be better for expressing geometric properties such as location.

\begin{figure}[t]
  \centering
  \includegraphics[width=0.5\textwidth]{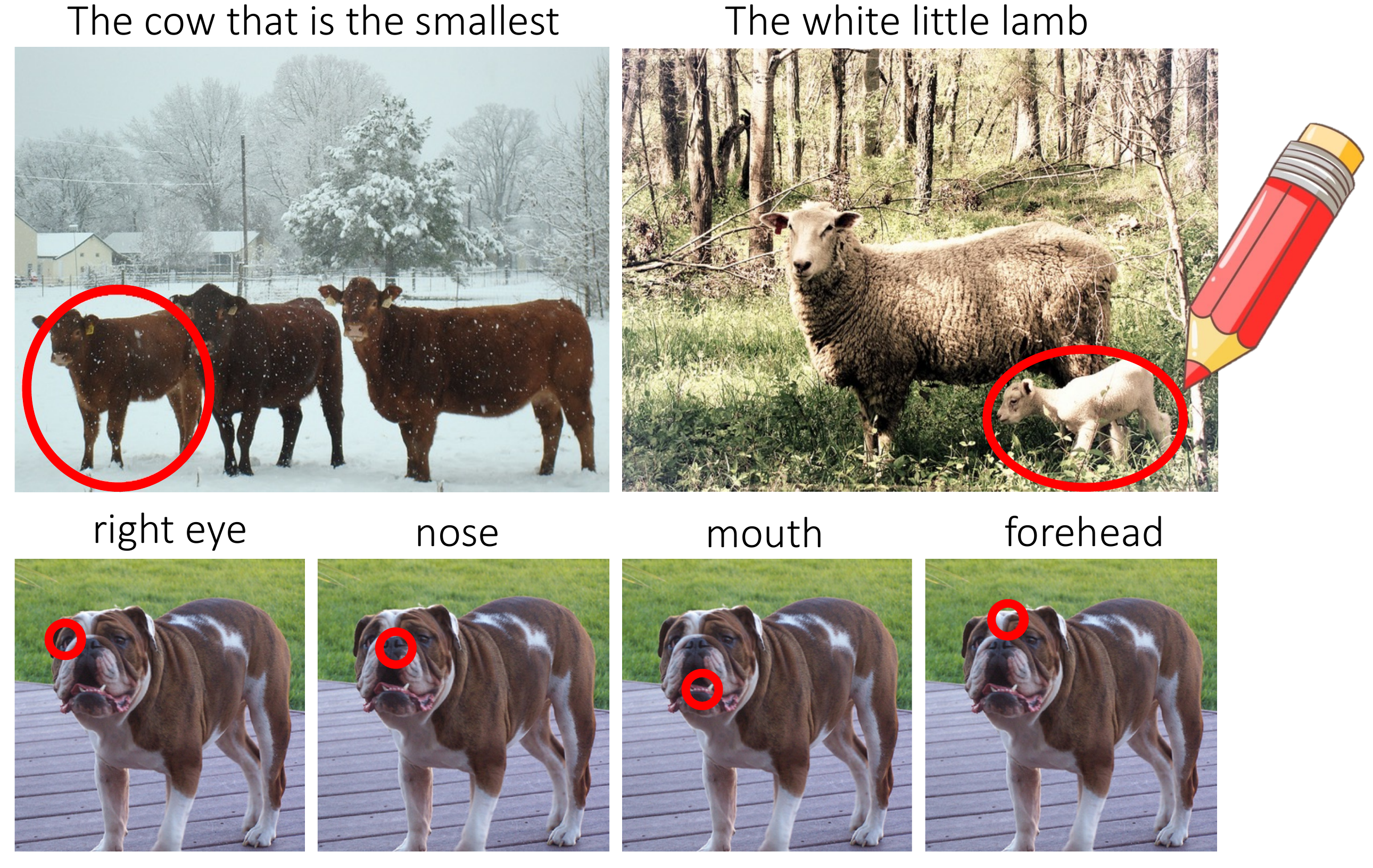}
  \caption{\textbf{Visual Prompt Engineering.} We draw multiple annotations over an image and have CLIP choose the correct one given a caption. Here we show predictions for the given expressions. Top: Examples from RefCOCOg on referring expressions detection. Bottom: Example from SPair71k on keypoint localization.}%
  \label{fig:teaser}
\end{figure}

In this paper, we thus explore \emph{visual prompt engineering}
\footnote{Note the difference between visual prompt \emph{tuning}, a setting previously explored, where the prompts are task-specific learnable tokens, and visual prompt \emph{engineering}, where we apply a fixed augmentation in pixel space.}.
We do so with two goals.
The first goal is to contribute one more practical tool for extracting useful information from VLMs in a zero-shot manner.
We demonstrate this by obtaining state-of-the-art zero-shot results in referring expressions comprehension by engineering visual prompts.
The second goal is to characterise interesting and unexpected properties of the VLMs and their training data, including identifying some behaviours that can raise ethical concerns.

Perhaps the most surprising of our findings is the effectiveness of a particular type of visual prompting: \textbf{drawing a plain red circle} on top of the image (\cref{fig:teaser}).
We show that this simple intervention steers the VLM to analyse/talk about the image region contained in the circle.
This behaviour can then be used for tasks such as naming a specific object or object part or detecting particular image regions based on a description.
The latter, for instance, is achieved by marking each object proposal with a red circle and using the VLM to find the best match with respect to the provided referring expression, achieving strong results on multiple benchmarks in the unsupervised regime.
Furthermore, we show that prompting with a circle also works for finer-grained localization, marking specific object parts or keypoints instead of just whole objects.

We further contrast marking an image with the  alternative of cropping it, which, from sliding window classifiers to region neural networks, is the canonical approach to steer the focus of an image-level predictor to a particular image region.
We show that, for VLMs at least, marking is significantly more effective than cropping, possibly because it does not lose contextual information like the latter.

Apart from the practical applications, our findings reveal unexpected and intriguing properties of VLMs.
We show empirically, that marking with a red circle is optimal among a selection of possible markers (variants of the circle, boxes, arrows, etc.).
Presumably, the VLMs understand red circles out of the box because these appear sufficiently frequently in the training corpus, i.e., the Internet.
While we do not have access to the full training data of CLIP, we corroborate this intuition by seeking examples of such images in YFCC15M, a dataset of CC-BY images.

Our analysis shows that red circles are indeed present even in a (comparatively small) dataset of images like YFCC15M, but they are \emph{rare}.
It is a testament to the extraordinary capacity of VLMs that such a behaviour can be learned from such rare events, without an explicit focus on doing so.
We test models of different sizes/capacities and show that only the larger models exhibit this behaviour reliably, which corresponds to our intuition.

Finally, we note that the ability of VLMs to learn even from rare events such ``red circles'' can acquire both desirable and undesirable behaviours.
Red circles, in particular, can have a negative connotation in the training data as they are often used by news outlets to mark missing people or criminals and, evidently, the model learns from such examples.
As a result, we show that drawing a red circle in an image increases the probability that the model would characterise a person as a criminal or as a missing person.

To summarise, we make the following main contributions:
(1) We propose marking as a new form of visual prompt engineering that is effective in extracting useful emergent behaviours in VLMs like CLIP;
(2) We use the latter to achieve state-of-the-art zero-shot referring expressions comprehension using a VLM;
(3) We provide an analysis of why marking is effective for these models, and link that to the training data and large model capacity;
(4) We show that visual prompt engineering can also elicit unwanted behaviours, such as triggering problematic biases in the VLMs, revealing potential ethical issues.
\section{Related work}%
\label{sec:related-work}

\textbf{Emergent Behaviour from Large Scale Pretraining} has mainly been observed in Large Language Models (LLMs).
Most notably, GPT-2~\cite{radford2019language-gpt2}, GPT-3~\cite{brown2020language_gpt3}, and ChatGPT~\cite{chatgpt} have been shown to be capable of tasks such as zero-shot translation, question answering, arithmetic, as well as planning actions for embodied agents~\cite{huang2022language-planning}.
Fine-tuning LLMs can also lead to models that can generate code from docstrings~\cite{chen2021evaluating-codex} or solve math problems~\cite{cobbe2021training-verifiers, lewkowycz2022solving}.
Only a few emergent zero-shot behaviours have been reported for VLMs like CLIP, mainly for classification~\cite{radford2021learning} and OCR~\cite{materzynska2022disentangling-clip}.
Generative VLMs like FLAMINGO~\cite{alayrac2022flamingo} and BLIP~\cite{li2023blip} excel in captioning and visual question-answering tasks, but also have no way of solving pixel-level computer vision tasks.

\textbf{Prompting VLMs} is most commonly performed by prepending a set of learnable tokens to the text input~\cite{guo2022texts, ju2022prompting-video, zhou2022conditional-prompt-learning, zhou2022learning-to-prompt}, vision input~\cite{jia2022visual_prompt_tuning, tu2022visual, zhang2022promptcal}, or both text and vision inputs~\cite{shen2022multitask, zang2022unified-prompting}, in order to easily steer a frozen CLIP model to solve a desired task. 
{}\cite{bahng2022visual-prompting} learn augmentations in pixel space, such as padding around the image, or changing a patch of the image, which are optimized with gradient descent on a downstream task.
{}\cite{bar2022visual} cast image inpainting as a visual prompting task, using a generative model trained on figures from academic papers.
Coloring regions of an image has been used for the VCR task~\cite{zellers2019recognition-vcr}, where a model is finetuned on annotated images~\cite{zellers2021merlot}.
Colorful Prompt Tuning (CPT)~\cite{yao2021cpt} color regions of an image and use a captioning model to predict which object in an image an expression refers to by predicting its color.
Similarly to CPT, we augment the input image in pixel space and perform zero-shot inference.
However, we \emph{annotate} the image in a human-like manner and show that our method is more powerful and more flexible than CPT.

\textbf{Referring Expression Comprehension} (REC)
aims to localize a target object in an image that corresponds to a textual description.
Most approaches to REC start with object proposals, for example, generated with Faster-RCNN~\cite{ren2015faster}, and learn to score them~\cite{hu2016natural, luo2017comprehension, liu2020learning_graph_rec, yang2019dynamic_graph_rec, wang2019neighbourhood_graph_rec}.
REC is sometimes considered together with referring expression generation --- the task of generating a description of a given region.
{}\cite{luo2017comprehension} use a comprehension model to guide a generator, whereas~\cite{chen2019referring_caption_aware} jointly train a detector with a caption generator.
Some works model the scene as a graph~\cite{liu2020learning_graph_rec, yang2019dynamic_graph_rec, wang2019neighbourhood_graph_rec}
or use language parsers and grammar-based methods~\cite{cirik2018using_grounding_rec, liu2019learning_grounding_rec}, leading to a more interpretable result.
More recently, transformer architectures have been used~\cite{li2021referring_transformer, kamath2021mdetr, dou2022coarse-fiber, yang2021crossing,wang2022cris}.
{}~\cite{li2021referring_transformer, kamath2021mdetr, dou2022coarse-fiber, yang2021crossing} perform text-modulated object detection, where a transformer decoder takes the referring expression as an input and predicts a bounding box.
{}~\cite{wang2022cris} train with a text-to-pixel contrastive loss, which allows for a text-driven segmentation or detection at test time.

\textbf{Unsupervised Referring Expression Comprehension} is a less explored area, only made possible with the introduction of large pre-trained models such as CLIP~\cite{radford2021learning}.
ReCLIP~\cite{subramanian2022reclip} crops object proposals and ranks them using CLIP before an ad-hoc postprocessing step to take into account relations such as left/right, smaller/bigger, etc.
CPT~\cite{yao2021cpt} colors object proposal boxes and use a pre-trained captioning model~\cite{zhang2021vinvl} to auto-regressively predict which colored proposal corresponds to the query description.
Pseudo-Q~\cite{jiang2022pseudoq} generates descriptions for multiple objects in an image, which is used to train a REC network.
However, this model is not fully unsupervised as the pseudo descriptions it uses are generated using a captioning model trained on COCO\@.

\textbf{Visual Reasoning Using Large Pretrained Models} has been an area of significant interest in the last few years.
In addition to referring expression detection~\cite{subramanian2022reclip}, CLIP~\cite{radford2021learning} has been used for semantic segmentation~\cite{pakhomov2021segmentation, liang2022open}.
{}\cite{pakhomov2021segmentation} use CLIP to assign text labels to object parts after doing part co-segmentation in the latent space of a GAN.
{}\cite{liang2022open} utilize CLIP for open-vocabulary segmentation by using a general-purpose mask proposal network and CLIP as a classifier.
CLIP has also been used for unsupervised object proposal generation~\cite{shi2022proposalclip} and open-set detection~\cite{esmaeilpour2021zero}.
Semantic segmentation also emerges from image only~\cite{melas2022deep, wang2022tokencut} or image-text~\cite{xu2022groupvit} self-supervision.

\textbf{Bias of VLMs} is an increasingly popular area of research, as downstream applications come with the risk of perpetuating biases and stereotypes existing in the training data.
However, methods for assessing the bias of a VLM are still not well established.
{}\cite{agarwal2021evaluating} measure the misclassification rate of CLIP of faces of people of different races with non-human and criminal categories, whereas~\cite{berg2022prompt, chuang2023debiasing, wang2021gender} measure fairness in retrieval results.
Here, we show  a different kind of bias, where the addition of a red circle over a person can trigger a negative connotation.
\begin{figure*}[t!]
\centering
\includegraphics[width=0.80\textwidth]{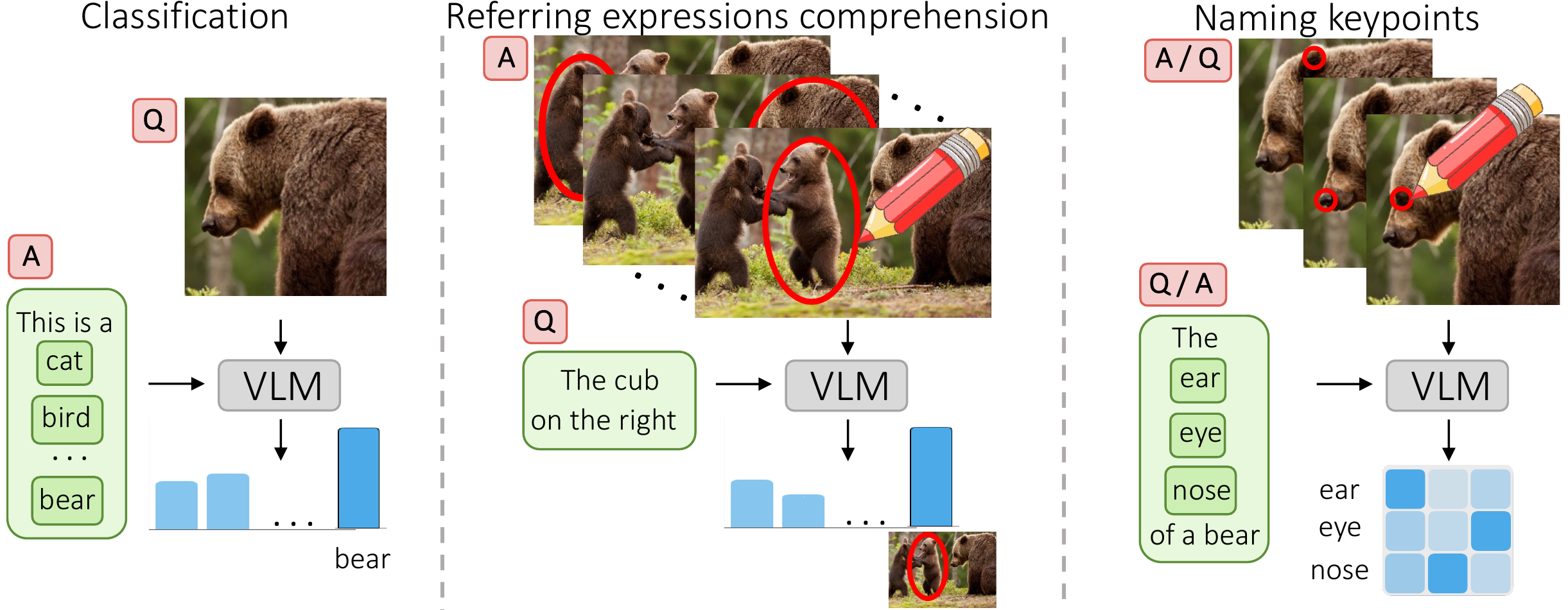}
\caption{\textbf{Prompt engineering for VLMs.} We cast zero-shot inference with VLMs as a Q/A problem, each requiring specific prompt engineering.
In the figure, Q is \emph{Question} and A is \emph{Answer} (a set of possible answers).
Left: text prompt engineering for classification.
This widely used method can be interpreted as follows in our framework:
The image is the question, and classes are the available answers, which are engineered into prompts.
Middle: visual prompt engineering for referring expressions comprehension.
The question is the referring expression, and the available answers are the box proposals, which we engineer into visual prompts.
Right: visual and text prompt engineering for keypoint matching.
For keypoint localization, we use a similar setup to referring expressions, where the question is a keypoint in plain text and the possible answers are all 2D locations in the image.}%
\label{fig:method}
\end{figure*}

\section{Method}%
\label{sec:method}

Our goal is to develop visual prompting in \emph{Vision-Language Models} (VLMs).
VLMs solve prediction tasks that involve jointly processing text and images.
For example, models such as CLIP are trained to \emph{match} text and image samples.
The input to such a VLM is an image $i \in \mathbb{R}^{3\times H \times W}$ and text $t \in \Sigma^*$, where $\Sigma$ is an alphabet.
The output is a score $s(i, t)$ that expresses the degree of compatibility between the supplied image and text.

\subsection{Prompt engineering}

One of the most striking capabilities of VLMs is their ability to solve a variety of classification tasks with little to no further training at all, in a zero-shot manner.
This is done by reducing the task of interest to that of evaluating the VLM on suitably-engineered image and text pairs.

For example, given an image-caption pair $(i,t)$, consider the problem of localizing a named object keypoint in the image.
We can cast this as a question-answer problem, where the question $q\in Q$ is the name of the object keypoint (e.g., ``right ear'', ``front left leg'', \dots) and the answer $a \in A$ is one of a discrete set of image locations.

Because the VLM computes a compatibility score $s(i,t)$ between an image $i$ and the text $t$, it cannot be used to map the question $q$ to the answer $a$ directly.
However, via prompt engineering, we can use the VLM to construct a compatibility score $s(q,a|i,t)$ between question and answer, conditioned on the input image-text pair $(i,t)$.
This score is in general given by the expression
\begin{equation}\label{e:general}
s(q,a|i,t) = s(i_{qa}, t_{qa}),
\end{equation}
where $i_{qa}$ and $t_{qa}$ are versions of the input image and text, obtained by transforming the latter to reflect the question-answer pair $(q,a)$.

The specific way \cref{e:general} should be applied to a problem depends on the specific nature of the latter.
For example, in the problem of localizing the named keypoints, it is natural to encode the name of the keypoint via the textual modality and its 2D location via the visual modality.
For instance, in order to answer the question $q=\text{``right ear''}$ for a given input image $i$ with caption $t=\text{``dog''}$, we can engineer the textual prompt
$
t_{qa} = t_q = \text{``an image of the right ear of a dog''}
$
to encode a description of the named entity.
Likewise, we can engineer the visual prompt
$
i_{qa} = i_a
$
in such a way as to `select' the location $a$ in the image, using one of the methods discussed in \cref{s:marking}.
With this, we can answer the question by finding
$
\hat a(q|i,t) = \operatornamewithlimits{argmax}_{a \in A} s(q,a|i,t),
$
that maximizes the score
$
s(q,a|i,t) = s(i_a, t_q),
$
which specializes \cref{e:general}.

In the following sections, we provide further details and apply these ideas to a few concrete tasks.

\subsection{Visual prompting via marking}\label{s:marking}

The usual way of encoding location information in a visual prompt is to crop the image around the desired location, meaning that $i_a$ is the image cropped around $a$.
This idea has been used extensively with VLMs, including to interpret referring expressions, where maximizing a score of the form $s(i_a,t_q)$ seeks for the image crop that best matches the referring expression $t_q$.

In this paper, we explore an alternative approach for visual prompting that uses the concept of \emph{marking} the desired region in the image.
Marking quite literally means overlaying to the image $i_a$ a circle, a box, or an arrow, which visually indicates the desired location $a$.

While the idea of marking may sound strange, it is interesting for two reasons.
First, differently from cropping, a marked image $i_a$ preserves almost all the information contained in the input image $i$, including contextual information that crops lack.
Second, we show that marking works well with VLMs, \emph{outperforming} cropping-based prompt engineering in some prediction tasks.

While the simplest marking consisting of a red circle is particularly effective, in \cref{sec:experiments} we explore several different ways of generating markings.
We refer the reader to that section for further details and examples.

\subsection{Tasks}%
\label{s:tasks}

We study the idea of mark-based prompt engineering by considering several zero-shot prediction tasks, from simple tasks such as matching keypoints to their names to more complex ones such as referring expression comprehension.

\paragraph{Naming Keypoints.}%
\label{s:matching}

The first and simplest task that we consider is matching the name of the keypoints of an object to their 2D locations in an image.
The input is an image $i$, a set of keypoint names $Q$, and a set of corresponding keypoint locations $A \subset \{0,\dots,H-1\}\times \{0,\dots,W-1\}$.
The number of names and locations is the same ($m=|Q|=|A|$) and the goal is to match the two.
We express the latter as predicting the square permutation matrix $\Pi \in S_m$ that associates each name $q$ to its corresponding location $a$ (i.e., $\Pi_{qa} = 1$).

In order to predict $\Pi$, we use \cref{e:general} to define the cost of associating name $q$ to location $a$ as
$
C_{qa} = s(i_{a}, t_q)
$
where $i_a$ is obtained either via cropping or marking and $t_q$ is just the name of the keypoints prefixed by the string ``an image of''.
For this problem, the role of questions and answers is symmetric and we decode the cost matrix $C$ into a permutation matrix $\Pi$ via optimal transport:
\begin{equation}
\hat\Pi(i,Q,A) = \operatornamewithlimits{argmax}_{\Pi\in S_m}
\sum_{q\in Q, a\in A}
\Pi_{qa} \exp\left(-\tau C_{qa}\right),
\end{equation}
where $\tau > 0$ is a temperature parameter.
This optimization problem is solved efficiently via the Sinkhorn-Knopp algorithm~\cite{sinkhorn1967concerning}, which renormalizes matrix $C$.

\paragraph{Keypoint Localization.}%
\label{s:kp-localization}

The second task is a more useful and difficult variant of the first.
The goal is still to localize a named keypoint $q$ in an image, but this time the locations $A$ are a subset of a $m\times m$ regular grid.
These are further restricted to a salient image region extracted by using the unsupervised saliency method of~\cite{wang2022tokencut} to avoid testing irrelevant locations in the background.
The difference compared to naming keypoints is that this version of the problem does not assume prior knowledge of the possible locations of the keypoints.
Given the name $q$ of a keypoint, its location $a$ is then obtained as
$
\hat a(i,q) = \operatornamewithlimits{argmax}_{a\in A} s(i_a,t_q)
$
where $i_a$ and $t_q$ are as defined previously.

\begin{table*}[th]
\centering
\resizebox{0.83\textwidth}{!}{%
\begin{tabular}{lccccccc|ccccccc}
\toprule
\multirow{3}{*}{Method} & \multicolumn{7}{c}{Name-to-keypoint} & \multicolumn{7}{c}{Keypoint-to-name} \\
& \multirow{2}{*}{\textbf{CUB}} & \multicolumn{6}{c}{\textbf{Spair71k}} & \multirow{2}{*}{\textbf{CUB}} & \multicolumn{6}{c}{\textbf{SPair71k}} \\
& & \faBird & \faCat & \faDog & \faHorse & \faSheep & \faCow & & \faBird & \faCat & \faDog & \faHorse & \faSheep & \faCow \\
\midrule
Random & 8.2 & 16.8 & 15.0 & 10.5 & 9.4 & 15.1 & 11.9 & 8.2 & 16.8 & 15.0 & 10.5 & 9.4 & 15.1 & 11.9\\
Crop w/o SK & 15.8 & 28.5 & 28.5 & 28.5 & 20.1 & 26.1 & 29.7 & 18.7 & 22.4 & 19.1 & 24.0 & 14.9 & 27.3 & 25.1 \\ 
Crop w/ SK & 25.5 & 35.1 & 37.5 & 34.6 & 23.9 & 32.9 & 36.3 & 25.8 & 36.1 & 32.5 & 32.7 & 19.8 & 35.3 & 32.5 \\ 
\midrule
\textbf{Red Circle w/o SK} & 46.5 & 54.8 & 53.1 & 51.6 & 40.1 & 47.4 & 45.2 & 29.5 & 26.8 & 24.9 & 36.9 & 18.8 & 31.8 & 28.9 \\ 
\textbf{Red Circle w/ SK} & \textbf{58.2} & \textbf{67.6} & \textbf{60.1} & \textbf{59.3} & \textbf{53.1} & \textbf{56.7} & \textbf{52.8} &  \textbf{56.5} & \textbf{67.2} & \textbf{54.4} & \textbf{59.7} & \textbf{49.8} & \textbf{56.6} & \textbf{53.0} \\
\bottomrule \\
\end{tabular}%
}%
\vspace{-1em}
\caption{\textbf{Naming keypoints results on CUB and SPair71k}. On SPair71k, we show results on all animal classes --- bird, cat, dog, horse, sheep, cow. We show the percentage of correctly matched keypoints and names, given a list of them. We compare to randomly guessing the correct correspondence and cropping around the region of interest, rather than drawing an annotation. We compare all methods with and without normalization with the Sinkhorn-Knopp (SK) algorithm.}%
\label{tab:results_clip_matching_all}
\end{table*}

\paragraph{Referring Expression Comprehension.}%
\label{s:referring-expression}

Comprehending a referring expression means detecting an object in an image that corresponds to a textual specification that explicitly refers to it (e.g., ``fourth dog from the right'').
Similarly to prior work~\cite{yao2021cpt, subramanian2022reclip, jiang2022pseudoq}, given an image $i$, we approach this problem by extracting first a set of object proposals using the method from~\cite{yu2018mattnet} and interpret those as the set of possible answers $A$.
The set of questions $Q$ is instead a collection of referring expressions extracted from a given benchmark dataset.
For each referring expression, the best matching proposal is then given by
\begin{equation}
\hat a(i,q) = \operatornamewithlimits{argmax}_{a\in A}
\left[
s(i_a,t_q)
-
\frac{1}{|Q|}
\sum_{\bar q \in Q} s(i_a,t_{\bar q})
\right].
\end{equation}

The engineered prompts $i_a$ and $t_q$ are defined as in \cref{s:matching}.
In this case, we found it useful to subtract from the score the average with respect to all possible referring expressions $Q$.
This weighs down hypotheses $a$ such as faces that are visually very salient and tend to respond very strongly to all questions $q$.

\section{Experiments}%
\label{sec:experiments}

We study the properties of visual marking in VLMs by considering first the three tasks of \cref{s:tasks}: naming keypoints, localizing keypoints, and referring expression comprehension.

\subsection{Naming Keypoints}\label{sec:kp}

Naming keypoints is a comparatively simple problem that has no direct application; however, it is simpler and faster to evaluate than the other tasks, so we use it to ablate various aspects of our method.

\paragraph{Data and implementation details.}

For this task, we consider the CUB-200-2011 (CUB)~\cite{welinder2010caltech} and SPair71k~\cite{min2019spair} datasets.
The first contains named keypoint annotations for each image, whereas the second only annotates matching keypoints in pairs of images, but does not name them.
We thus augment the latter, manually naming each keypoint instance in each animal image.
We further crop the images from SPair71k with the provided bounding boxes.
For the VLM, we use the ViT-L/14@336px backbone.
Please see the sup.~matt.~for details.

\paragraph{Results.}

\begin{figure}[t]
\centering
\includegraphics[width=0.38\textwidth]{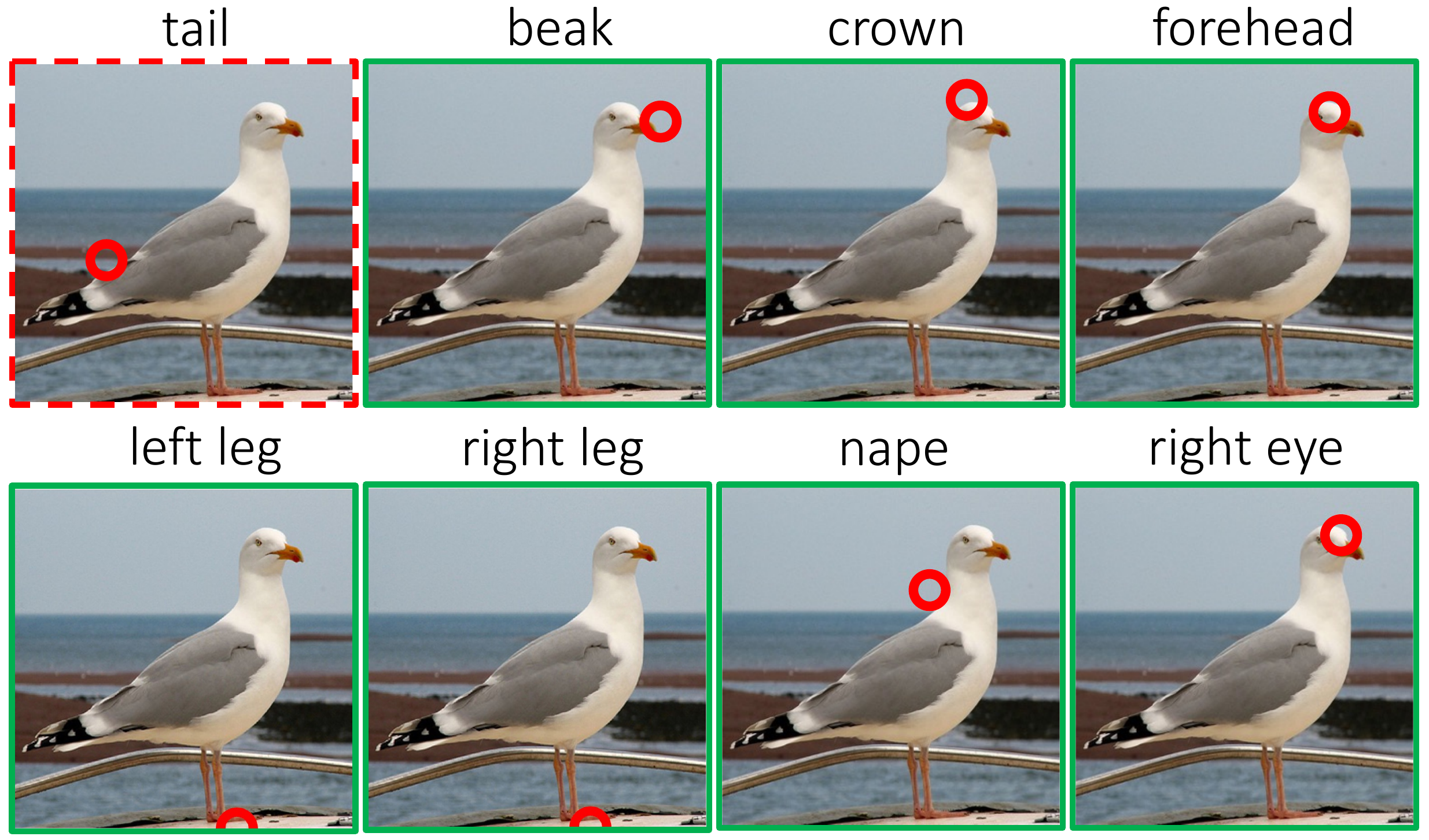}
\caption{\textbf{Qualitative Results on Localizing Keypoints} on an image from SPair71k. 
Green and red (dashed) borders are for correct and wrong predictions according to PCK with $\alpha=0.1$.
The red circle shown is thicker than the one used;
see the sup.~matt.~for examples of the actual thickness.}%
\label{fig:result_localization}
\end{figure}

Recall that, in this task, the output of the predictor is a permutation matrix $\Pi$ associating each keypoint location to a corresponding name.
We report (i) the ratio of keypoint names that are mapped to the correct location and (ii) the ratio of keypoint locations that are mapped to the correct names.
To the best of our knowledge, there are no prior works that associate keypoints with their names.
We thus compare the result of this new task to (a) random choice and (b) a baseline where $i_a$ is obtained by cropping.

As seen in \cref{tab:results_clip_matching_all}, prompting via visual marking (red circles) significantly outperforms the baselines, achieving almost twice the accuracy.
Using the Sinkhorn-Knopp (SK) algorithm to normalize the matching score further boosts results, mainly improving
results for points that are ambiguous and close to each other, \eg, mouth and nose.

\paragraph{What is the best visual marker?}

\begin{figure}[t]
\centering
\includegraphics[width=0.45\textwidth]{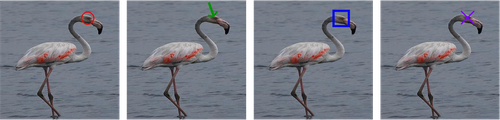}
\caption{\textbf{Example Annotations for Keypoints.} We show how the annotations we use look for different shapes and colors. We also experiment with multiple thicknesses and sizes. The red circle annotation on the left is the one we use throughout all evaluations unless stated otherwise.}%
\label{fig:example-annos-kps}
\end{figure}

We compare the use of (i) different shapes for highlighting a location: circle, rectangle, cross, arrow, (ii) different sizes, and (iii) different colors of the annotations, and show some examples in~\cref{fig:example-annos-kps}.
We compare different shapes and colors in~\cref{tab:keypoint_naming_shape_color_comparison} and find that red circles perform best.
Red is the best color despite the fact that it is a commonly occurring color in images, unlike colors like purple which can be found less often in nature and can thus be more distinctive, but lead to worse performance.
We attribute this to the fact that this emergent capability of CLIP exists due to human-centric manipulations of its training data, and humans are likely to annotate using red circles, as shown next.

\begin{table}[t]
\centering
\begin{minipage}[t]{0.22\textwidth}
\centering
\resizebox{\textwidth}{!}{%
\begin{tabular}{cccc}
\toprule
Marker shape & Mean & Best \\
\midrule
\textbf{Circle} & \textbf{33.5 $\pm$ 4.5} & \textbf{46.5} \\
Arrow & 28.3 $\pm$ 3.1 & 36.3 \\
Square & 24.1 $\pm$ 3.6 & 36.3 \\
Cross & 21.5 $\pm$ 6.3 & 34.5\\
\bottomrule
\end{tabular}%
}
\end{minipage}
\hfill
\begin{minipage}[t]{0.22\textwidth}
\centering
\resizebox{\textwidth}{!}{%
\begin{tabular}{ccc}
\toprule
Circle color & Mean & Best \\
\midrule
\textbf{Red} & \textbf{36.4 $\pm$ 5.1} & \textbf{46.5} \\
Green & 34.3 $\pm$ 4.2 & 43.3 \\
Purple & 34.0 $\pm$ 3.7 & 41.9 \\
Blue & 32.7 $\pm$ 3.9 & 41.1 \\
Yellow & 32.4 $\pm$ 4.0 & 40.8 \\

\bottomrule \\
\end{tabular}%
}
\end{minipage}
\caption{\textbf{Ablation of annotation types for naming keypoint.} 
We evaluate on CUB and present results across a variety of colors and sizes (for marker shape) and sizes (for circle color). 
For full results refer to the sup.~matt.}%
\label{tab:keypoint_naming_shape_color_comparison}
\end{table}

\paragraph{Are there visual markers in the training data?}

\begin{figure}[t]
\centering
\includegraphics[width=0.45\textwidth]{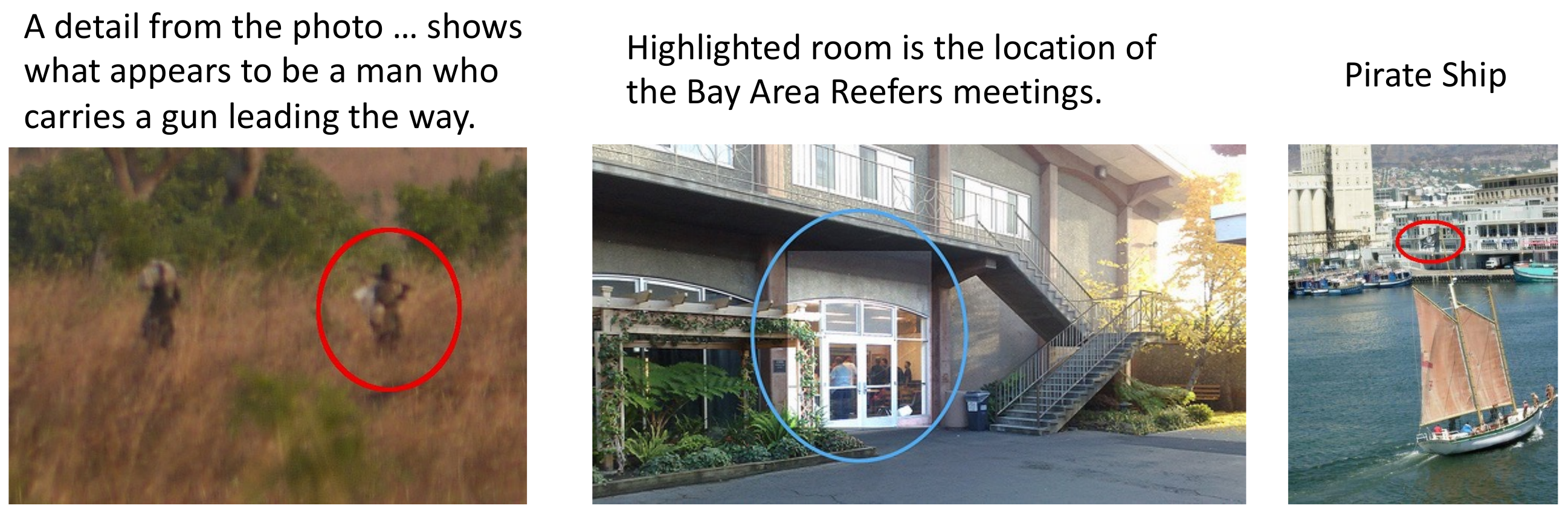}
\caption{\textbf{Discovered annotations in YFCC15M.} We show images from YFCC15M that have been human-annotated, and their corresponding captions. Discovered by training a simple detector.}%
\label{fig:annotated-images-yfcc15m}
\end{figure}

To explore the hypothesis that CLIP can zero-shot classify annotations on images because of similar examples seen during training, we find images in YFCC15M that contain markers
(YFCC15M is a subset of the CLIP training data).
To this end, we train a binary classifier using an ensemble of a ViT-B/16 and RN50x16 CLIP vision encoders to classify images in YFCC15M that contain annotations.
We then use this to filter a 6M subset of YFCC15M and take the top 10k images with the highest score.
Finally, we manually examine the 10k images and find 70 images that have annotations drawn on top of them.
We show 3 such images in \cref{fig:annotated-images-yfcc15m}.
Hence, the training data contains examples of markers, but they are very rare ($\sim$0.001\%), suggesting that such behavior can only be learned from very large datasets by high-capacity models.
This is further explored next.

\paragraph{How do different VLMs differ?}

\begin{table}[]
\centering
\resizebox{0.45\textwidth}{!}{%
\begin{tabular}{llllcc|cc}
\toprule
\multirow{2}{*}{Method} &\multirow{2}{*}{Backbone} & \multirow{2}{*}{Data} & \multirow{2}{*}{Params} &\multicolumn{2}{c}{Name-to-keypoint} & \multicolumn{2}{c}{Keypoint-to-name} \\
& & & & \textbf{CUB} & \textbf{Spair71k} & \textbf{CUB} & \textbf{SPair71k} \\

\midrule
CLIP~\cite{radford2021learning} & ViT-B/32 & 400M & 87M & 19.1 & 26.7 & 19.1 & 25.6 \\
CLIP~\cite{radford2021learning} & ViT-B/16 & 400M & 86M &22.2 & 34.0 & 22.1 & 33.6 \\
CLIP~\cite{radford2021learning} & RN50x16 & 400M & 167M & 30.6 & 41.0 & 29.7 & 40.0 \\
CLIP~\cite{radford2021learning} & ViT-L/14 & 400M & 304M & 47.9 & 54.3 & 48.0 & 51.2 \\
CLIP~\cite{radford2021learning} & ViT-L/14$\star$ & 400M & 304M & \textbf{58.2} & \textbf{58.3} & \textbf{56.5} & \textbf{56.8} \\
\midrule
OpenCLIP~\cite{cherti2022openclip} & ViT-B/32 & 2B & 87M & 19.4 & 27.5 & 20.7 & 27.2\\
OpenCLIP~\cite{cherti2022openclip} & ViT-L/14 & 2B & 304M & 33.9 & 42.4 & 33.3 & 41.5 \\
OpenCLIP~\cite{cherti2022openclip} & ViT-H/14 & 2B & 632M & 45.0 & 53.7 & 42.8 & 50.5\\
OpenCLIP~\cite{cherti2022openclip} & ViT-g/14 & 2B & 1.01B & 44.2 & 47.2 & 42.5 & 43.6\\
OpenCLIP~\cite{cherti2022openclip} & ViT-G/14 & 2B & 1.84B & 50.4 & 52.5 & 48.9 & 48.6\\
\midrule
SLIP~\cite{mu2022slip} & ViT-S/16 & 15M & 22M & 13.0 & 17.8 & 12.0 & 16.7 \\
SLIP~\cite{mu2022slip} & ViT-B/16 & 15M & 86M & 17.3 & 16.5 & 16.7 & 17.1 \\
SLIP~\cite{mu2022slip} & ViT-L/16 & 15M & 303M & 24.6 & 26.3 & 24.1 & 25.0 \\
CLIP$\dagger$~\cite{mu2022slip, radford2021learning}& ViT-S/16 & 15M & 22M & 11.4 & 16.5 & 12.4 & 15.8 \\
CLIP$\dagger$~\cite{mu2022slip, radford2021learning} & ViT-B/16 & 15M & 86M & 14.0 & 18.4 & 14.8 & 17.8 \\
CLIP$\dagger$~\cite{mu2022slip, radford2021learning} & ViT-L/16 & 15M & 303M & 15.0 & 20.5 & 15.8 & 20.4 \\
\midrule
FILIP$\ddagger$~\cite{li2021declip, yao2021filip}& ViT-B/32 & 15M & 90M & 8.9 & 15.6 & 8.7 & 15.6 \\
DeFILIP~\cite{li2021declip} & ViT-B/32 & 15M & 90M & 12.5 & 19.5 & 12.5 & 19.4 \\
CLIP$\ddagger$~\cite{li2021declip, radford2021learning} & ViT-B/32 & 15M & 90M & 10.6 & 14.1 & 11.0 & 14.3 \\
DeCLIP~\cite{li2021declip} & ViT-B/32 & 15M & 90M & 15.8 & 19.9 & 15.8 & 18.7 \\
DeCLIP~\cite{li2021declip} & ViT-B/32 & 88M & 90M & 19.4 & 23.6 & 19.7 & 22.3 \\

\bottomrule \\
\end{tabular}
}
\caption{\textbf{Comparison of backbones and pretraining methods}. We compare several VL pretraining methods, implementations, and backbones. $\dagger$ and $\ddagger$ : we evaluate the models trained by~\cite{mu2022slip} and~\cite{li2021declip}, respectively. $\star$ : we evaluate a model trained on images of input size $336 \times 336$. For SPair71k, we present the average across all animal classes. The 400M, 2B, 15M, and 88M datasets are WIT-400M, LAION-2B, YFCC-15M and DeCLIP-88M, respectively.}%
\label{tab:results_clip_matching_compare_models}
\end{table}

\begin{figure}[t]
\centering
\includegraphics[width=0.44\textwidth]{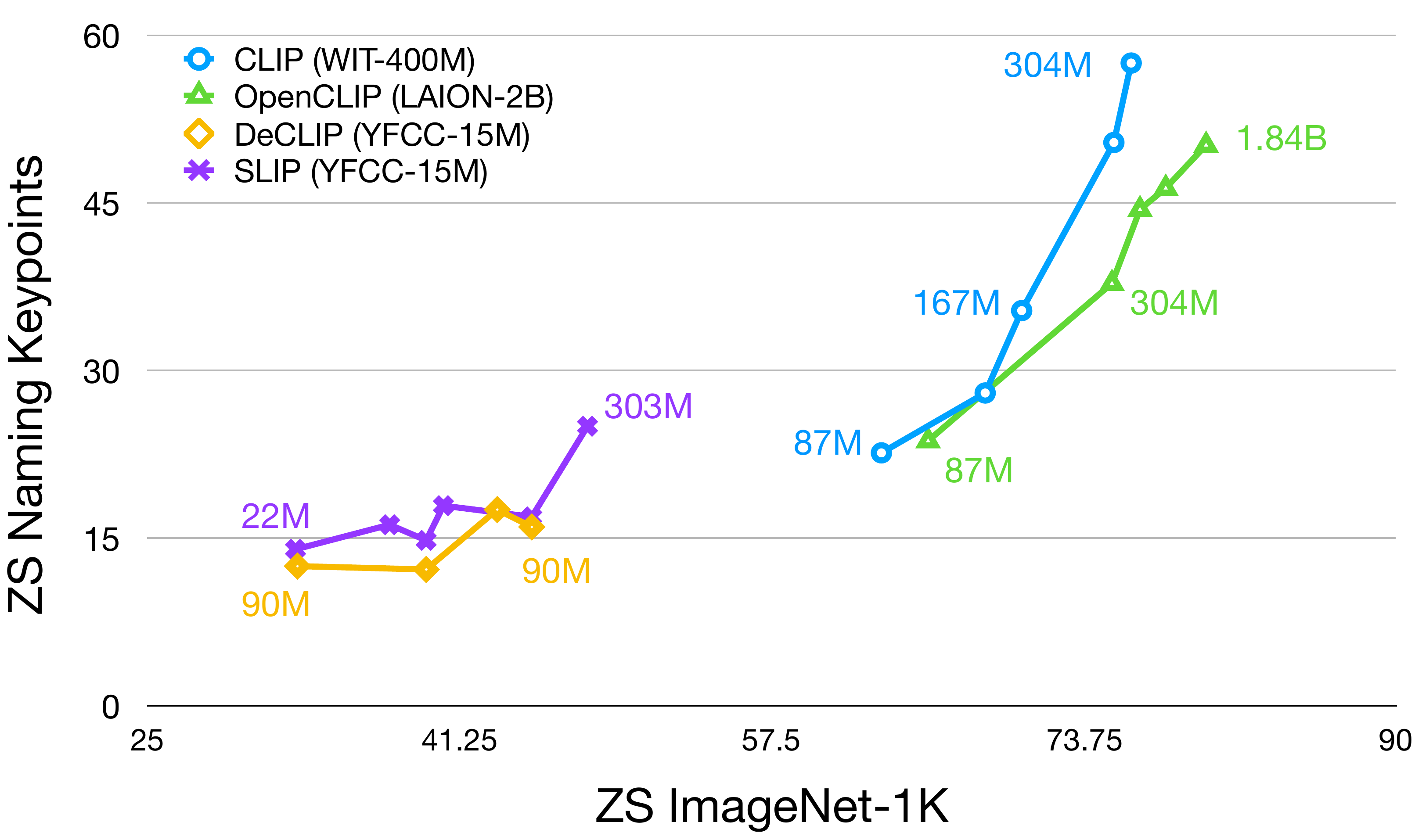}
\caption{\textbf{Comparison of Zero-shot (ZS) Naming Keypoints to ZS ImageNet-1K.} We show results for CLIP~\cite{radford2021learning}, OpenCLIP~\cite{cherti2022openclip}, DeCLIP~\cite{li2021declip} and SLIP~\cite{mu2022slip}. We compute the average text-to-image and image-to-text matching score on SPair71k and CUB and compare it against the reported zeros-shot ImageNet-1K accuracy for each model. For all methods, we show the pretraining dataset and the number of parameters of the vision encoders of some of the models.}%
\label{fig:matching_vs_imagenet}
\end{figure}

We compare a number of CLIP models in \cref{fig:matching_vs_imagenet}.
In general, we observe that the performance of keypoint matching improves with (i) the size of the pretraining dataset and (ii) the size of the vision encoder.
The former holds true for CLIP models trained on WIT-400M vs YFCC-15M (which is a subset of WIT-400M).
However, using the LAION-2B dataset for pretraining leads to worse results.
We suspect this result comes from differences in filtering when creating WIT-400M and LAION-2B, where in the latter, examples of annotations might have been discarded due to a stronger focus on \textit{aesthetic} images for generative models.
Similarly, we see big gains in performance as we increase the size of the vision encoder, and the gains do not seem to converge with the biggest available models.
We emphasize on the dramatic increase in performance of the WIT-400M pretrained CLIP --- the biggest models improve on the performance of the smallest by 250\% on ZS keypoint matching, whereas the improvement on ZS ImageNet-1K classification is just 20\%.
We draw similarities between this task and tasks in the domain of NLP, such as zero-shot or one-shot arithmetic, where only the largest GPT models perform well~\cite{brown2020language_gpt3}.
We argue that in a similar fashion, the vision encoder needs sufficient capacity and data in order to show this emergent behavior.

\subsection{Localizing Keypoints}\label{sec:localkp}

\begin{table}[t]
\centering
\resizebox{0.42\textwidth}{!}{%
\begin{tabular}{lcccccccc}
\toprule
\multirow{2}{*}{Method} & \multirow{2}{*}{Mask} & \multirow{2}{*}{\textbf{CUB}} & \multicolumn{6}{c}{\textbf{Spair71k}}  \\
& & & \faBird & \faCat & \faDog & \faHorse & \faSheep & \faCow \\
\midrule
Random & \xmark & 1.1 & 3.1 & 1.8 & 3.6 & 3.0 & 3.0 & 3.4  \\
Random & \cmark & 8.3 & 6.7 & 4.7 & 4.6 & 5.5 & 3.3 &  4.3 \\
Crop & \xmark & 16.9 & 25.5 & 27.3 & 22.6 & 16.1 & 20.0 & 22.7  \\
Crop & \cmark & 21.3 & 28.4 & 28.8 & 21.9 & 16.0 & 23.4 & 24.5  \\
\midrule
\textbf{Red Circle} & \xmark & 32.3 & 50.7 & 52.7 & 55.9 & 38.3 & 42.4 & 38.0  \\
\textbf{Red Circle} & \cmark & \textbf{45.2} & \textbf{53.4} & \textbf{54.5} & \textbf{56.4} & \textbf{40.9} & \textbf{43.2} & \textbf{42.2}  \\
\bottomrule \\
\end{tabular}%
}
\vspace{-1em}
\caption{\textbf{Named Keypoint Localization Results.} We report PCK of predicted keypoint location and compare to a baseline where we crop around the region.}%
\label{tab:results_localization}
\end{table}

For this experiment, we use the same data and network architecture as for the previous one, but report the \emph{percentage of correct keypoints} (PCK) as a metric, as the latter is  widely used when evaluating semantic correspondences.
Given a set of ground-truth points $\mathcal{P}=\{p_{m}\}^{M}_{m=1}$ and predictions $\mathcal{\hat{P}}=\{p_{m}\}^{M}_{m=1}$, PCK is given by:
\begin{equation*}
\mathrm{PCK}(\mathcal{P}, \mathcal{\hat{P}})= \frac{1}{M}\sum^{M}_{m=1} \mathbbm{1}\lbrack \| \hat{p}_{m}-p_{m} \| \leq \delta \rbrack .
\end{equation*}
Here, $\delta$ is a distance threshold given by $\delta = \alpha \max(H, W)$, where $0<\alpha<1$ is a ratio and $(H, W)$ is the bounding box size.
For all datasets we use $\alpha=0.1$.
Keypoint localization also utilizes an unsupervised saliency mask to ignore background locations.

Similarly to the naming task, we compare keypoint localization to random guessing and the crop-based baseline.
As shown in \cref{tab:results_localization}, using red circles significantly outperforms both;
as expected, results are further improved by using saliency to further filter keypoint locations.
We show qualitative results in~\cref{fig:result_localization}.

\subsection{Referring Expression Comprehension}\label{sec:refexp}

\newcommand{\g}[1]{\textcolor{gray}{#1}}
\begin{table}
\begin{center}
\resizebox{0.45\textwidth}{!}{%
\begin{tabular}{lcccc|ccc|cc}
    \toprule
\multirow{2}{*}{Method} & \multirow{2}{*}{ZS} &\multicolumn{3}{c|}{\textbf{RefCOCO}} & \multicolumn{3}{c|}{\textbf{RefCOCO+}} & \multicolumn{2}{c}{\textbf{RefCOCOg}} \\
 & & Val & TestA & TestB & Val & TestA & TestB & Val & Test \\
    \midrule
 \g{DTWREG~\cite{sun2021discriminative}}  & \g{\xmark}   & \g{39.2} & \g{41.1} & \g{37.7} & \g{39.2} & \g{40.1} & \g{38.1} & \g{--} & \g{--} \\
\g{Pseudo-Q {}\cite{jiang2022pseudoq}} & \g{\xmark} & \g{56.0} & \g{58.3} & \g{54.1} & \g{38.9} & \g{45.1} & \g{32.1} & \g{46.3} & \g{47.4} \\
\midrule
CPT {}\cite{yao2021cpt} & \cmark & 32.2 & 36.1 & 30.3 & 31.9 & 35.2 & 28.8 &  36.7 & 36.5 \\
ReCLIP $\dagger$ {}\cite{subramanian2022reclip} & \cmark & 42.0 & 43.5 & 39.0 & 47.4 & 50.1 & 43.9 & 57.8 & 57.2 \\
ReCLIP $\ddagger$ {}\cite{subramanian2022reclip} & \cmark & 45.8 & 46.1 & \textbf{47.1} & 47.9 & 50.1 & 45.1 & 59.3 & \textbf{59.0} \\
\textbf{Red Circle} & \cmark & \textbf{49.8} & \textbf{58.6} & 39.9 & \textbf{55.3} & \textbf{63.9} & \textbf{45.4} & \textbf{59.4} & 58.9 \\
    \bottomrule \\
\end{tabular}}
\end{center}
\vspace{-2em}
\caption{\textbf{Comparison with state-of-the-art on REC.} We report top-1 accuracy (\%).
$\dagger$ is the crop-based baseline of~\cite{subramanian2022reclip} and $\ddagger$ is their method that adds relational resolving.
\textit{``ZS''} refers to zero-shot approaches.
Drawing red circles outperforms other zero-shot approaches on most benchmarks, including ReCLIP$\ddagger$ that post-processes results using manually designed relational rules.
On RefCOCO+ and RefCOCOg, a red circle also outperforms Pseudo-Q and DTWREG that are not zero-shot and use weak supervision. }%
\label{tab:ref-exp-results-sota}
\end{table}

\paragraph{Datasets and implementation details.}

Referring expression comprehension is commonly evaluated on the RefCOCO~\cite{yu2016modeling}, RefCOCO+~\cite{yu2016modeling}, and RefCOCOg~\cite{mao2016generation} datasets, all of which consist of images from the MS-COCO dataset~\cite{lin2014microsoft} together with  expressions that refer to a unique object in the image, which are also annotated with a bounding box.
RefCOCO+ only contains appearance-based expressions, whereas RefCOCO and RefCOCOg contain relation-based expressions (e.g., containing the words left/closer/bigger).
The test sets of RefCOCO and RefCOCO+ are split in two, where ``testA'' and ``testB'' contain only people and non-people, respectively.
We evaluate using the percentage of correct predictions, where a box is correctly predicted if its intersection-over-union with the ground-truth box is over 0.5.

For the referring expressions task, we use an ensemble RN50x16 and ViT-L/14@336 CLIP backbones.
Following prior work~\cite{subramanian2022reclip, yao2021cpt}, we score the bounding box proposals of MAttNet~\cite{yu2018mattnet}.

\paragraph{Results.}
\newcommand{\red}[1]{\textcolor{red}{#1}}
\newcommand{\gr}[1]{\textcolor{ForestGreen}{#1}}

\begin{table*}[t]
\centering
\resizebox{0.75\textwidth}{!}{%
\begin{tabular}{lcccc|ccc}
\toprule
\multirow{2}{*}{Model} & \multirow{2}{*}{Red Circle} & \multicolumn{3}{c|}{\textbf{FaceSynthetics}} & \multicolumn{3}{c}{\textbf{COCO}} \\

& & Positive & Neutral & Criminal & Positive & Neutral & Criminal \\
 
\midrule
ViT-L/14@336px & \xmark & 0.5\% & 35.6\% & 63.9\% & 22.7\% & 47.5\% & 29.8\%\\
ViT-L/14@336px & \cmark & 0.0\%  & 19.1\% & 80.9\% \red{(+17.0\%)} & 1.6\% & 505.\% & 47.9 \red{(+18.1\%)}\\
\midrule
ViT-L/14 & \xmark & 1.3\% & 40.8\% & 57.9\% & 25.0\% & 46.5\% & 28.6\%\\
ViT-L/14 & \cmark & 0.0\% & 32.5\% & 67.5\% \red{(+9.6\%)} & 2.7\% & 43.1\% & 54.2 \red{(+25.6\%)}\\
\midrule
ViT-B/16 & \xmark & 0.7\% & 49.1\% & 50.2\% & 19.4\% & 56.7\% & 23.9\%\\
ViT-B/16 & \cmark & 0.0\% & 37.1\% & 62.9\ \red{(+12.7\%)} & 4.7\% & 39.3\% & 56.0 \red{(+32.1\%)}\\
\midrule
ViT-B/32 & \xmark & 0.0\% & 85.2\% & 14.8\%  & 15.3\% & 61.8\% & 22.9\%\\
ViT-B/32 & \cmark & 0.0\% & 70.5\% & 29.5\% \red{(+14.7\%)}& 4.9\% & 48.2\% & 46.9 \red{(+24.0\%)}\\
\midrule
RN50x16 & \xmark & 0.7\% & 46.4\% & 52.9\%  & 26.5\% & 30.5\% & 43.0\%\\
RN50x16 & \cmark & 0.1\% & 52.2\% & 47.7\% \gr{(-5.2\%)} & 13.6\% & 39.5\% & 46.9 \red{(+3.9\%)}\\
\bottomrule \\
\end{tabular}%
}%
\vspace{-1em}
\caption{\textbf{Classification rate into criminal categories.} We report ZS classification results in criminal categories on synthetic faces from FaceSynthetics and persons from COCO, with and without red circles. In brackets is the absolute difference from the crop without a red circle.}%
\label{tab:bias_eval}
\end{table*}
\begin{figure}[t]
\centering
\includegraphics[width=0.45\textwidth]{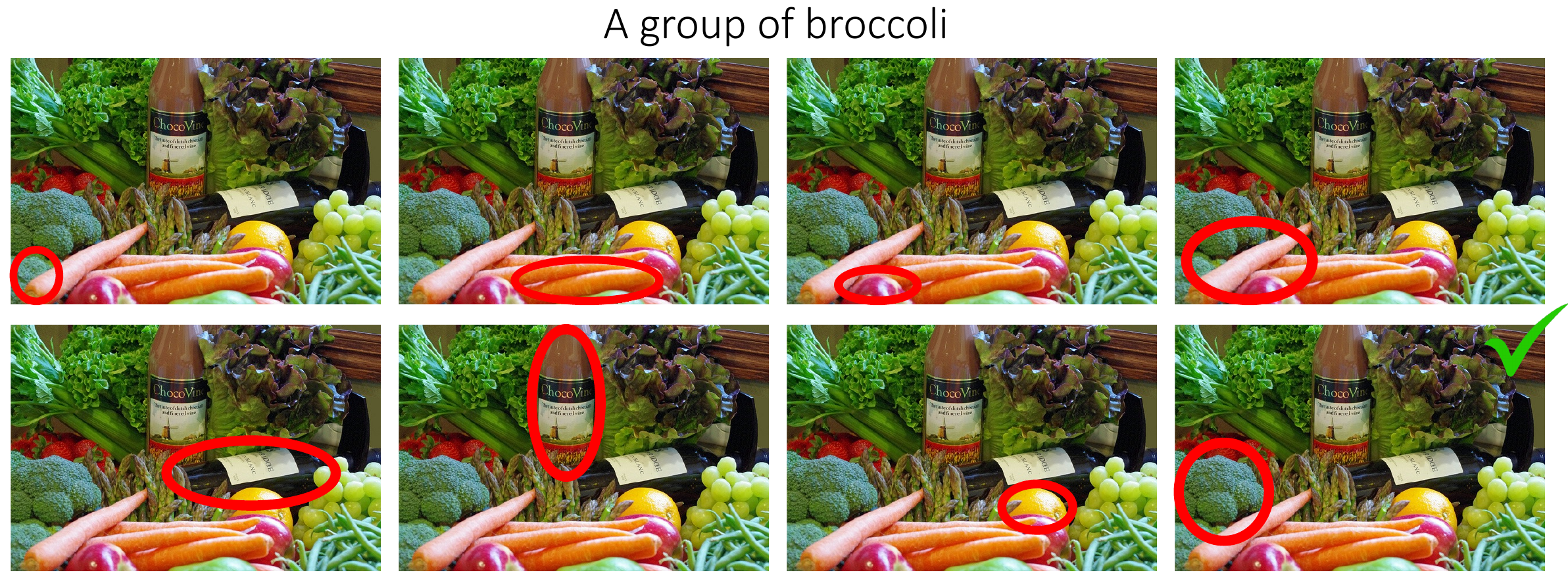}
\caption{\textbf{Qualitative Evaluation on REC} on an image from RefCOCOg.
CLIP correctly chooses the bottom right image.
The circle is shown thicker than actually used for clarity (see the sup.~matt.~for the actual thickness).}%
\label{fig:results_rec}
\end{figure}

Using a red circle, we achieve state-of-the-art on most referring expressions comprehension baselines in the zero-shot setting, as shown in \cref{tab:ref-exp-results-sota}.
Interestingly, this even outperforms ReCLIP~\cite{subramanian2022reclip}, which is based on scoring image crops, followed by post-processing with manually designed relations rules.
A red circle also outperforms Pseudo-Q on most benchmarks, even though Pseudo-Q explicitly trains for this task.

\subsection{Model biases and ethics}\label{sec:bias}

While drawing circles on images can extract useful behaviors from a VLM for a wide variety of legitimate image analysis tasks, it can also extract unwanted ones and \textbf{must not be used for the analysis of sensitive data}.

\begin{figure}[t]
\centering
\includegraphics[width=0.48\textwidth]{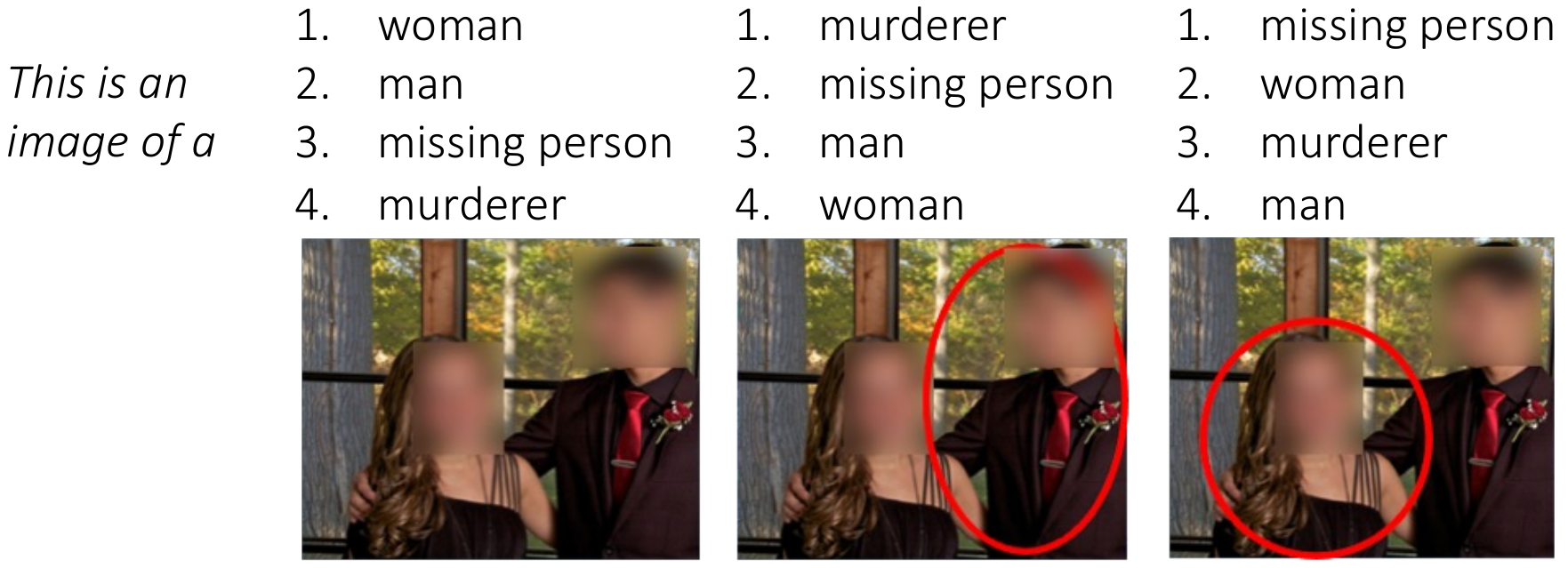}%
\caption{\textbf{Bias of CLIP.}
We zero-shot classify a random COCO image with a male-looking and a female-looking individual, also adding a circle over each of them. We score the following sentences to the images:
\textit{This is an image of a \{woman, man, missing person, suspected murderer \}.}
The apparent gender resolution is correct, but the circled images tend to be scored higher as missing or murderers.
Blur added for privacy.}%
\label{fig:biases}
\end{figure}

To demonstrate this fact, in \cref{fig:biases} we take a random image from COCO that contains a male-looking and a female-looking individual and zero-shot classify the image, as well as the image with circles over each individual, into 4 categories: male, female, missing person, and suspected murderer.
While this leads to a correct resolution of the apparent gender of the annotated person, the annotated images are more likely to be classified as containing a missing person or a murderer.
While we cannot know for certain, we hypothesize that this is due to the presence in the CLIP model training data of missing person reports, police footage, or similar, where people have been marked.

We further quantify these biases following~\cite{agarwal2021evaluating}, using synthetic faces from FaceSynthetics~\cite{wood2021fake} and person crops from COCO~\cite{lin2014microsoft}.
For FaceSyntetics, we take 1000 random synthetic faces, and for COCO, we crop all bounding boxes for the class \emph{person} from the validation set that have an area of at least 10\% of the total area of the image, which comes down to 1352 crops.
Following~\cite{agarwal2021evaluating}, we measure zero-shot classification rates into criminal categories. We introduce a ``positive'' category (honest man/woman/person), ``neutral'' category (man/woman/person) and a ``criminal'' category (criminal/thief/suspicious person).
Finally, we zero-shot classify the original images and the images with circles.
In \cref{tab:bias_eval} we present classification rates into criminal categories.
We see that for all ViT encoders, the rate at which people are classified as criminals is significantly higher.
This is problematic as such existing biases can lead to harmful consequences.
Note that there are various limitations in this analysis, including the usage of binary gender.

\NewDocumentCommand\emojione{}{\scalerel*{\includegraphics{1F44F}}{X}}

\section{Conclusions}%
\label{sec:conclusions}

We have shown that visual prompt engineering via marking can extract useful behavior from VLMs such as CLIP in a zero-shot manner, achieving state-of-the-art zero-shot referring expression comprehension performance, and significantly outperforming traditional techniques like image cropping.
Our analysis suggests that this behavior emerges because relevant samples of marking exist in the training data of the VLMs, but these samples are very rare.
As a consequence, the behavior can only be learned by very large models trained on very large datasets.
The analysis also shows that VLMs acquire undesirable behaviors too, where the mere addition of a red circle to an image increases the model's belief that the image has a negative connotation.

\paragraph{Dataset Ethics.}

We use the RefCOCO, RefCOCO+, MS-COCO, FaceSynthetics, YFCC15M, CUB, SPair71k in a manner compatible with their terms.
Some of these images may contain personal data (faces). In \cref{sec:kp,sec:localkp,sec:refexp} there is no extraction of biometric data.
In \cref{sec:bias} we use MS-COCO to demonstrate that such a method cannot reliably extract information about people due to the bias in the pre-trained CLIP model (there is no identification).
The FaceSynthetics, used for the same purpose, is a dataset of synthetic faces, so it does not raise privacy concerns.
For further details on ethics, data protection, and copyright please see \url{https://www.robots.ox.ac.uk/~vedaldi/research/union/ethics.html}.

\paragraph{Acknowledgements.}
We thank Luke Melas-Kyriazi, Tim Franzmeyer, Rhydian Windsor and Bruno Korbar for proofreading
A.~Shtedritski is supported by EPSRC EP/S024050/1.
A.~Vedaldi and C.~Rupprecht are supported by ERC-CoG UNION 101001212.
C.~Rupprecht is also partially supported by VisualAI EP/T028572/1.

{\small\bibliographystyle{ieee_fullname}
\bibliography{refs}}

\begin{thebibliography}{10}\itemsep=-1pt

\bibitem{chatgpt}
Chatgpt.
\newblock \url{https://chat.openai.com/}.

\bibitem{agarwal2021evaluating}
Sandhini Agarwal, Gretchen Krueger, Jack Clark, Alec Radford, Jong~Wook Kim,
  and Miles Brundage.
\newblock Evaluating clip: towards characterization of broader capabilities and
  downstream implications.
\newblock {\em arXiv preprint arXiv:2108.02818}, 2021.

\bibitem{alayrac2022flamingo}
Jean-Baptiste Alayrac, Jeff Donahue, Pauline Luc, Antoine Miech, Iain Barr,
  Yana Hasson, Karel Lenc, Arthur Mensch, Katie Millican, Malcolm Reynolds,
  et~al.
\newblock Flamingo: a visual language model for few-shot learning.
\newblock {\em arXiv preprint arXiv:2204.14198}, 2022.

\bibitem{bahng2022visual-prompting}
Hyojin Bahng, Ali Jahanian, Swami Sankaranarayanan, and Phillip Isola.
\newblock Visual prompting: Modifying pixel space to adapt pre-trained models.
\newblock {\em arXiv preprint arXiv:2203.17274}, 2022.

\bibitem{bar2022visual}
Amir Bar, Yossi Gandelsman, Trevor Darrell, Amir Globerson, and Alexei~A Efros.
\newblock Visual prompting via image inpainting.
\newblock {\em arXiv preprint arXiv:2209.00647}, 2022.

\bibitem{berg2022prompt}
Hugo Berg, Siobhan~Mackenzie Hall, Yash Bhalgat, Wonsuk Yang, Hannah~Rose Kirk,
  Aleksandar Shtedritski, and Max Bain.
\newblock A prompt array keeps the bias away: Debiasing vision-language models
  with adversarial learning.
\newblock {\em arXiv preprint arXiv:2203.11933}, 2022.

\bibitem{brown2020language_gpt3}
Tom Brown, Benjamin Mann, Nick Ryder, Melanie Subbiah, Jared~D Kaplan, Prafulla
  Dhariwal, Arvind Neelakantan, Pranav Shyam, Girish Sastry, Amanda Askell,
  et~al.
\newblock Language models are few-shot learners.
\newblock {\em Advances in neural information processing systems},
  33:1877--1901, 2020.

\bibitem{chen2021evaluating-codex}
Mark Chen, Jerry Tworek, Heewoo Jun, Qiming Yuan, Henrique Ponde de~Oliveira
  Pinto, Jared Kaplan, Harri Edwards, Yuri Burda, Nicholas Joseph, Greg
  Brockman, et~al.
\newblock Evaluating large language models trained on code.
\newblock {\em arXiv preprint arXiv:2107.03374}, 2021.

\bibitem{chen2019referring_caption_aware}
Yi-Wen Chen, Yi-Hsuan Tsai, Tiantian Wang, Yen-Yu Lin, and Ming-Hsuan Yang.
\newblock Referring expression object segmentation with caption-aware
  consistency.
\newblock {\em arXiv preprint arXiv:1910.04748}, 2019.

\bibitem{cherti2022openclip}
Mehdi Cherti, Romain Beaumont, Ross Wightman, Mitchell Wortsman, Gabriel
  Ilharco, Cade Gordon, Christoph Schuhmann, Ludwig Schmidt, and Jenia Jitsev.
\newblock Reproducible scaling laws for contrastive language-image learning.
\newblock {\em arXiv preprint arXiv:2212.07143}, 2022.

\bibitem{chuang2023debiasing}
Ching-Yao Chuang, Varun Jampani, Yuanzhen Li, Antonio Torralba, and Stefanie
  Jegelka.
\newblock Debiasing vision-language models via biased prompts.
\newblock {\em arXiv preprint arXiv:2302.00070}, 2023.

\bibitem{cirik2018using_grounding_rec}
Volkan Cirik, Taylor Berg-Kirkpatrick, and Louis-Philippe Morency.
\newblock Using syntax to ground referring expressions in natural images.
\newblock In {\em Proceedings of the AAAI conference on artificial
  intelligence}, volume~32, 2018.

\bibitem{cobbe2021training-verifiers}
Karl Cobbe, Vineet Kosaraju, Mohammad Bavarian, Mark Chen, Heewoo Jun, Lukasz
  Kaiser, Matthias Plappert, Jerry Tworek, Jacob Hilton, Reiichiro Nakano,
  et~al.
\newblock Training verifiers to solve math word problems.
\newblock {\em arXiv preprint arXiv:2110.14168}, 2021.

\bibitem{dou2022coarse-fiber}
Zi-Yi Dou, Aishwarya Kamath, Zhe Gan, Pengchuan Zhang, Jianfeng Wang, Linjie
  Li, Zicheng Liu, Ce Liu, Yann LeCun, Nanyun Peng, et~al.
\newblock Coarse-to-fine vision-language pre-training with fusion in the
  backbone.
\newblock {\em arXiv preprint arXiv:2206.07643}, 2022.

\bibitem{esmaeilpour2021zero}
Sepideh Esmaeilpour, Bing Liu, Eric Robertson, and Lei Shu.
\newblock Zero-shot open set detection by extending clip.
\newblock {\em arXiv preprint arXiv:2109.02748}, 2021.

\bibitem{guo2022texts}
Zixian Guo, Bowen Dong, Zhilong Ji, Jinfeng Bai, Yiwen Guo, and Wangmeng Zuo.
\newblock Texts as images in prompt tuning for multi-label image recognition.
\newblock {\em arXiv preprint arXiv:2211.12739}, 2022.

\bibitem{hu2016natural}
Ronghang Hu, Huazhe Xu, Marcus Rohrbach, Jiashi Feng, Kate Saenko, and Trevor
  Darrell.
\newblock Natural language object retrieval.
\newblock In {\em Proceedings of the IEEE conference on computer vision and
  pattern recognition}, pages 4555--4564, 2016.

\bibitem{huang2022language-planning}
Wenlong Huang, Pieter Abbeel, Deepak Pathak, and Igor Mordatch.
\newblock Language models as zero-shot planners: Extracting actionable
  knowledge for embodied agents.
\newblock In {\em International Conference on Machine Learning}, pages
  9118--9147. PMLR, 2022.

\bibitem{jia2022visual_prompt_tuning}
Menglin Jia, Luming Tang, Bor-Chun Chen, Claire Cardie, Serge Belongie, Bharath
  Hariharan, and Ser-Nam Lim.
\newblock Visual prompt tuning.
\newblock {\em arXiv preprint arXiv:2203.12119}, 2022.

\bibitem{jiang2022pseudoq}
Haojun Jiang, Yuanze Lin, Dongchen Han, Shiji Song, and Gao Huang.
\newblock Pseudo-q: Generating pseudo language queries for visual grounding.
\newblock In {\em Proceedings of the IEEE/CVF Conference on Computer Vision and
  Pattern Recognition}, pages 15513--15523, 2022.

\bibitem{ju2022prompting-video}
Chen Ju, Tengda Han, Kunhao Zheng, Ya Zhang, and Weidi Xie.
\newblock Prompting visual-language models for efficient video understanding.
\newblock In {\em Computer Vision--ECCV 2022: 17th European Conference, Tel
  Aviv, Israel, October 23--27, 2022, Proceedings, Part XXXV}, pages 105--124.
  Springer, 2022.

\bibitem{kamath2021mdetr}
Aishwarya Kamath, Mannat Singh, Yann LeCun, Gabriel Synnaeve, Ishan Misra, and
  Nicolas Carion.
\newblock Mdetr-modulated detection for end-to-end multi-modal understanding.
\newblock In {\em Proceedings of the IEEE/CVF International Conference on
  Computer Vision}, pages 1780--1790, 2021.

\bibitem{lewkowycz2022solving}
Aitor Lewkowycz, Anders Andreassen, David Dohan, Ethan Dyer, Henryk
  Michalewski, Vinay Ramasesh, Ambrose Slone, Cem Anil, Imanol Schlag, Theo
  Gutman-Solo, et~al.
\newblock Solving quantitative reasoning problems with language models.
\newblock {\em arXiv preprint arXiv:2206.14858}, 2022.

\bibitem{li2023blip}
Junnan Li, Dongxu Li, Silvio Savarese, and Steven Hoi.
\newblock Blip-2: Bootstrapping language-image pre-training with frozen image
  encoders and large language models.
\newblock {\em arXiv preprint arXiv:2301.12597}, 2023.

\bibitem{li2021referring_transformer}
Muchen Li and Leonid Sigal.
\newblock Referring transformer: A one-step approach to multi-task visual
  grounding.
\newblock {\em Advances in Neural Information Processing Systems},
  34:19652--19664, 2021.

\bibitem{li2021declip}
Yangguang Li, Feng Liang, Lichen Zhao, Yufeng Cui, Wanli Ouyang, Jing Shao,
  Fengwei Yu, and Junjie Yan.
\newblock Supervision exists everywhere: A data efficient contrastive
  language-image pre-training paradigm.
\newblock {\em arXiv preprint arXiv:2110.05208}, 2021.

\bibitem{liang2022open}
Feng Liang, Bichen Wu, Xiaoliang Dai, Kunpeng Li, Yinan Zhao, Hang Zhang,
  Peizhao Zhang, Peter Vajda, and Diana Marculescu.
\newblock Open-vocabulary semantic segmentation with mask-adapted clip.
\newblock {\em arXiv preprint arXiv:2210.04150}, 2022.

\bibitem{lin2014microsoft}
Tsung-Yi Lin, Michael Maire, Serge Belongie, James Hays, Pietro Perona, Deva
  Ramanan, Piotr Doll{\'a}r, and C~Lawrence Zitnick.
\newblock Microsoft coco: Common objects in context.
\newblock In {\em Computer Vision--ECCV 2014: 13th European Conference, Zurich,
  Switzerland, September 6-12, 2014, Proceedings, Part V 13}, pages 740--755.
  Springer, 2014.

\bibitem{liu2019learning_grounding_rec}
Daqing Liu, Hanwang Zhang, Feng Wu, and Zheng-Jun Zha.
\newblock Learning to assemble neural module tree networks for visual
  grounding.
\newblock In {\em Proceedings of the IEEE/CVF International Conference on
  Computer Vision}, pages 4673--4682, 2019.

\bibitem{liu2020learning_graph_rec}
Yongfei Liu, Bo Wan, Xiaodan Zhu, and Xuming He.
\newblock Learning cross-modal context graph for visual grounding.
\newblock In {\em Proceedings of the AAAI Conference on Artificial
  Intelligence}, volume~34, pages 11645--11652, 2020.

\bibitem{luo2017comprehension}
Ruotian Luo and Gregory Shakhnarovich.
\newblock Comprehension-guided referring expressions.
\newblock In {\em Proceedings of the IEEE Conference on Computer Vision and
  Pattern Recognition}, pages 7102--7111, 2017.

\bibitem{mao2016generation}
Junhua Mao, Jonathan Huang, Alexander Toshev, Oana Camburu, Alan~L Yuille, and
  Kevin Murphy.
\newblock Generation and comprehension of unambiguous object descriptions.
\newblock In {\em Proceedings of the IEEE conference on computer vision and
  pattern recognition}, pages 11--20, 2016.

\bibitem{materzynska2022disentangling-clip}
Joanna Materzy{\'n}ska, Antonio Torralba, and David Bau.
\newblock Disentangling visual and written concepts in clip.
\newblock In {\em Proceedings of the IEEE/CVF Conference on Computer Vision and
  Pattern Recognition}, pages 16410--16419, 2022.

\bibitem{melas2022deep}
Luke Melas-Kyriazi, Christian Rupprecht, Iro Laina, and Andrea Vedaldi.
\newblock Deep spectral methods: A surprisingly strong baseline for
  unsupervised semantic segmentation and localization.
\newblock In {\em Proceedings of the IEEE/CVF Conference on Computer Vision and
  Pattern Recognition}, pages 8364--8375, 2022.

\bibitem{min2019spair}
Juhong Min, Jongmin Lee, Jean Ponce, and Minsu Cho.
\newblock Spair-71k: A large-scale benchmark for semantic correspondence.
\newblock {\em arXiv preprint arXiv:1908.10543}, 2019.

\bibitem{mu2022slip}
Norman Mu, Alexander Kirillov, David Wagner, and Saining Xie.
\newblock Slip: Self-supervision meets language-image pre-training.
\newblock In {\em Computer Vision--ECCV 2022: 17th European Conference, Tel
  Aviv, Israel, October 23--27, 2022, Proceedings, Part XXVI}, pages 529--544.
  Springer, 2022.

\bibitem{pakhomov2021segmentation}
Daniil Pakhomov, Sanchit Hira, Narayani Wagle, Kemar~E Green, and Nassir Navab.
\newblock Segmentation in style: Unsupervised semantic image segmentation with
  stylegan and clip.
\newblock {\em arXiv preprint arXiv:2107.12518}, 2021.

\bibitem{radford2021learning}
Alec Radford, Jong~Wook Kim, Chris Hallacy, Aditya Ramesh, Gabriel Goh,
  Sandhini Agarwal, Girish Sastry, Amanda Askell, Pamela Mishkin, Jack Clark,
  et~al.
\newblock Learning transferable visual models from natural language
  supervision.
\newblock In {\em International Conference on Machine Learning}, pages
  8748--8763. PMLR, 2021.

\bibitem{radford2019language-gpt2}
Alec Radford, Jeffrey Wu, Rewon Child, David Luan, Dario Amodei, Ilya
  Sutskever, et~al.
\newblock Language models are unsupervised multitask learners.
\newblock {\em OpenAI blog}, 1(8):9, 2019.

\bibitem{ren2015faster}
Shaoqing Ren, Kaiming He, Ross Girshick, and Jian Sun.
\newblock Faster r-cnn: Towards real-time object detection with region proposal
  networks.
\newblock {\em Advances in neural information processing systems}, 28, 2015.

\bibitem{shen2022multitask}
Sheng Shen, Shijia Yang, Tianjun Zhang, Bohan Zhai, Joseph~E Gonzalez, Kurt
  Keutzer, and Trevor Darrell.
\newblock Multitask vision-language prompt tuning.
\newblock {\em arXiv preprint arXiv:2211.11720}, 2022.

\bibitem{shi2022proposalclip}
Hengcan Shi, Munawar Hayat, Yicheng Wu, and Jianfei Cai.
\newblock Proposalclip: Unsupervised open-category object proposal generation
  via exploiting clip cues.
\newblock In {\em Proceedings of the IEEE/CVF Conference on Computer Vision and
  Pattern Recognition}, pages 9611--9620, 2022.

\bibitem{sinkhorn1967concerning}
Richard Sinkhorn and Paul Knopp.
\newblock Concerning nonnegative matrices and doubly stochastic matrices.
\newblock {\em Pacific Journal of Mathematics}, 21(2):343--348, 1967.

\bibitem{subramanian2022reclip}
Sanjay Subramanian, Will Merrill, Trevor Darrell, Matt Gardner, Sameer Singh,
  and Anna Rohrbach.
\newblock Reclip: A strong zero-shot baseline for referring expression
  comprehension.
\newblock {\em arXiv preprint arXiv:2204.05991}, 2022.

\bibitem{sun2021discriminative}
Mingjie Sun, Jimin Xiao, Eng~Gee Lim, Si Liu, and John~Y Goulermas.
\newblock Discriminative triad matching and reconstruction for weakly referring
  expression grounding.
\newblock {\em IEEE transactions on pattern analysis and machine intelligence},
  43(11):4189--4195, 2021.

\bibitem{tu2022visual}
Cheng-Hao Tu, Zheda Mai, and Wei-Lun Chao.
\newblock Visual query tuning: Towards effective usage of intermediate
  representations for parameter and memory efficient transfer learning.
\newblock {\em arXiv preprint arXiv:2212.03220}, 2022.

\bibitem{wang2021gender}
Jialu Wang, Yang Liu, and Xin~Eric Wang.
\newblock Are gender-neutral queries really gender-neutral? mitigating gender
  bias in image search.
\newblock {\em arXiv preprint arXiv:2109.05433}, 2021.

\bibitem{wang2019neighbourhood_graph_rec}
Peng Wang, Qi Wu, Jiewei Cao, Chunhua Shen, Lianli Gao, and Anton van~den
  Hengel.
\newblock Neighbourhood watch: Referring expression comprehension via
  language-guided graph attention networks.
\newblock In {\em Proceedings of the IEEE/CVF Conference on Computer Vision and
  Pattern Recognition}, pages 1960--1968, 2019.

\bibitem{wang2022tokencut}
Yangtao Wang, Xi Shen, Yuan Yuan, Yuming Du, Maomao Li, Shell~Xu Hu, James~L
  Crowley, and Dominique Vaufreydaz.
\newblock Tokencut: Segmenting objects in images and videos with
  self-supervised transformer and normalized cut.
\newblock {\em arXiv preprint arXiv:2209.00383}, 2022.

\bibitem{wang2022cris}
Zhaoqing Wang, Yu Lu, Qiang Li, Xunqiang Tao, Yandong Guo, Mingming Gong, and
  Tongliang Liu.
\newblock Cris: Clip-driven referring image segmentation.
\newblock In {\em Proceedings of the IEEE/CVF Conference on Computer Vision and
  Pattern Recognition}, pages 11686--11695, 2022.

\bibitem{welinder2010caltech}
Peter Welinder, Steve Branson, Takeshi Mita, Catherine Wah, Florian Schroff,
  Serge Belongie, and Pietro Perona.
\newblock Caltech-ucsd birds 200.
\newblock 2010.

\bibitem{wood2021fake}
Erroll Wood, Tadas Baltru\v{s}aitis, Charlie Hewitt, Sebastian Dziadzio,
  Matthew Johnson, Virginia Estellers, Thomas~J. Cashman, and Jamie Shotton.
\newblock Fake it till you make it: Face analysis in the wild using synthetic
  data alone, 2021.

\bibitem{xu2022groupvit}
Jiarui Xu, Shalini De~Mello, Sifei Liu, Wonmin Byeon, Thomas Breuel, Jan Kautz,
  and Xiaolong Wang.
\newblock Groupvit: Semantic segmentation emerges from text supervision.
\newblock In {\em Proceedings of the IEEE/CVF Conference on Computer Vision and
  Pattern Recognition}, pages 18134--18144, 2022.

\bibitem{yang2019dynamic_graph_rec}
Sibei Yang, Guanbin Li, and Yizhou Yu.
\newblock Dynamic graph attention for referring expression comprehension.
\newblock In {\em Proceedings of the IEEE/CVF International Conference on
  Computer Vision}, pages 4644--4653, 2019.

\bibitem{yang2021crossing}
Zhengyuan Yang, Zhe Gan, Jianfeng Wang, Xiaowei Hu, Faisal Ahmed, Zicheng Liu,
  Yumao Lu, and Lijuan Wang.
\newblock Crossing the format boundary of text and boxes: Towards unified
  vision-language modeling.
\newblock {\em arXiv preprint arXiv:2111.12085}, 2021.

\bibitem{yao2021filip}
Lewei Yao, Runhui Huang, Lu Hou, Guansong Lu, Minzhe Niu, Hang Xu, Xiaodan
  Liang, Zhenguo Li, Xin Jiang, and Chunjing Xu.
\newblock Filip: fine-grained interactive language-image pre-training.
\newblock {\em arXiv preprint arXiv:2111.07783}, 2021.

\bibitem{yao2021cpt}
Yuan Yao, Ao Zhang, Zhengyan Zhang, Zhiyuan Liu, Tat-Seng Chua, and Maosong
  Sun.
\newblock Cpt: Colorful prompt tuning for pre-trained vision-language models.
\newblock {\em arXiv preprint arXiv:2109.11797}, 2021.

\bibitem{yu2018mattnet}
Licheng Yu, Zhe Lin, Xiaohui Shen, Jimei Yang, Xin Lu, Mohit Bansal, and
  Tamara~L Berg.
\newblock Mattnet: Modular attention network for referring expression
  comprehension.
\newblock In {\em Proceedings of the IEEE Conference on Computer Vision and
  Pattern Recognition}, pages 1307--1315, 2018.

\bibitem{yu2016modeling}
Licheng Yu, Patrick Poirson, Shan Yang, Alexander~C Berg, and Tamara~L Berg.
\newblock Modeling context in referring expressions.
\newblock In {\em Computer Vision--ECCV 2016: 14th European Conference,
  Amsterdam, The Netherlands, October 11-14, 2016, Proceedings, Part II 14},
  pages 69--85. Springer, 2016.

\bibitem{zang2022unified-prompting}
Yuhang Zang, Wei Li, Kaiyang Zhou, Chen Huang, and Chen~Change Loy.
\newblock Unified vision and language prompt learning.
\newblock {\em arXiv preprint arXiv:2210.07225}, 2022.

\bibitem{zellers2019recognition-vcr}
Rowan Zellers, Yonatan Bisk, Ali Farhadi, and Yejin Choi.
\newblock From recognition to cognition: Visual commonsense reasoning.
\newblock In {\em Proceedings of the IEEE/CVF conference on computer vision and
  pattern recognition}, pages 6720--6731, 2019.

\bibitem{zellers2021merlot}
Rowan Zellers, Ximing Lu, Jack Hessel, Youngjae Yu, Jae~Sung Park, Jize Cao,
  Ali Farhadi, and Yejin Choi.
\newblock Merlot: Multimodal neural script knowledge models.
\newblock {\em Advances in Neural Information Processing Systems},
  34:23634--23651, 2021.

\bibitem{zhang2021vinvl}
Pengchuan Zhang, Xiujun Li, Xiaowei Hu, Jianwei Yang, Lei Zhang, Lijuan Wang,
  Yejin Choi, and Jianfeng Gao.
\newblock Vinvl: Revisiting visual representations in vision-language models.
\newblock In {\em Proceedings of the IEEE/CVF Conference on Computer Vision and
  Pattern Recognition}, pages 5579--5588, 2021.

\bibitem{zhang2022promptcal}
Sheng Zhang, Salman Khan, Zhiqiang Shen, Muzammal Naseer, Guangyi Chen, and
  Fahad Khan.
\newblock Promptcal: Contrastive affinity learning via auxiliary prompts for
  generalized novel category discovery.
\newblock {\em arXiv preprint arXiv:2212.05590}, 2022.

\bibitem{zhou2022conditional-prompt-learning}
Kaiyang Zhou, Jingkang Yang, Chen~Change Loy, and Ziwei Liu.
\newblock Conditional prompt learning for vision-language models.
\newblock In {\em Proceedings of the IEEE/CVF Conference on Computer Vision and
  Pattern Recognition}, pages 16816--16825, 2022.

\bibitem{zhou2022learning-to-prompt}
Kaiyang Zhou, Jingkang Yang, Chen~Change Loy, and Ziwei Liu.
\newblock Learning to prompt for vision-language models.
\newblock {\em International Journal of Computer Vision}, 130(9):2337--2348,
  2022.

\end{thebibliography}

\clearpage
\begin{center}
	\textbf{\Large Appendix}
\end{center}

\begin{table*}
\begin{center}
\begin{tabular}{llccc|ccc|cc}
\toprule
\multirow{2}{*}{Method} & \multirow{2}{*}{Backbone} &\multicolumn{3}{c|}{\textbf{RefCOCO}} & \multicolumn{3}{c|}{\textbf{RefCOCO+}} & \multicolumn{2}{c}{\textbf{RefCOCOg}} \\
& & Val & TestA & TestB & Val & TestA & TestB & Val & Test \\
\midrule

\multirow{15}{*}{\textbf{ReCLIP}} & RN50$\times$16 & 37.61 & 38.32 & 37.19 & 44.12 & 46.02 & 41.81  & 55.94 & 54.36\\
&ViT-B/32 & 40.69 & 43.98 & 37.55 & 45.00 & 48.15 & 41.65 & 55.25 & 54.35\\
&ViT-B/16 & 38.23 & 40.53 & 37.00 & 41.53 & 42.91 & 41.32 & 55.19 & 55.16\\
&ViT-L/14 & 34.40 & 33.52 & 34.35 & 37.86 & 37.53 & 37.70 & 53.82 & 52.25 \\
&ViT-L/14@336px & 35.90 & 37.72 & 35.66 & 40.06 & 42.49 & 39.07 & 54.25 & 53.92\\
&RN50$\times$16,ViT-B/32 & \textbf{41.96} & \underline{43.52} & \underline{39.00} & \textbf{47.44} & \textbf{50.11} & \textbf{43.93} & 57.76 & \textbf{57.15}\\
&RN50$\times$16,ViT-B/1 & 39.94 & 41.61 & 38.71 & 45.06 & 47.17 & \underline{43.63} & \underline{57.93} & 56.85\\
&RN50$\times$16,ViT-L/14 & 37.98 & 38.08 & 37.51 & 42.87 & 44.57 & 41.66 & 56.78 & 56.02\\
&RN50$\times$16,ViT-L/14@336px & 38.79 & 39.49 & 37.82 & 44.27 & 46.44 & 42.46 & 57.86 & 56.28\\
&ViT-B/32,ViT-B/16 & \underline{41.34} & \textbf{44.25} & 38.55 & \underline{45.20} & 48.01 & 43.36 & 57.37 & 56.52 \\
&ViT-B/32,ViT-L/14 & 39.68 & 41.65 & 37.84 & 43.74 & 46.25 & 41.17 & 56.74 & 56.07\\
&ViT-B/32,ViT-L/14@336px & 40.82 & 43.47 & \textbf{39.22} & 45.41 & \underline{48.52} & 42.83 & \textbf{58.09} & \underline{56.94}\\
&ViT-B/16,ViT-L/14 & 37.69 & 38.29 & 37.53 & 40.87 & 42.07 & 40.93 & 56.35 & 55.76\\
& ViT-B/16,ViT-L/14@336px & 39.18 & 41.01 & 38.35 & 42.81 & 44.32 & 42.07 & 57.82 & 56.21\\
& ViT-L/14,ViT-L/14@336px & 35.47 & 36.26 & 35.70 & 39.52 & 40.69 & 38.70 & 54.51 & 54.04\\
\midrule
\multirow{15}{*}{\textbf{Red Circle}} & RN50$\times$16 & 45.52 & 52.99 & 38.59 & 49.98 & 57.55 & 42.11 & 53.94 & 54.35\\
& ViT-B/32 & 38.72 & 45.09 & 33.52 & 42.85 & 49.46 & 36.53 & 45.81 & 45.57\\
& ViT-B/16 & 45.30 & 52.70 & 36.51 & 49.39 & 57.67 & 40.60 & 53.72 & 53.26\\
& ViT-L/14 & 46.71 & 55.03 & 39.24 & 52.07 & 58.63 & 42.83 & 57.00 & 56.40\\
& ViT-L/14@336 & 48.27 & 56.44 & 39.71 & 53.59 & 59.99 & 43.28 & \textbf{59.95} & 58.51\\
& RN50$\times$16, ViT-B/32 & 45.62 & 54.04 & 37.13 & 50.73 & 60.46 & 41.69 & 54.00 & 53.84\\
& RN50$\times$16, ViT-B/16 &  \textbf{49.98} & 57.15 & 38.04 & 52.98 & 61.95 & 42.99 & 56.01 & 55.78\\
& RN50$\times$16, ViT-L/14 & 48.50 & 58.03 & 39.76 & 54.56 & 63.17 & \underline{44.41} & 58.17 & 57.76\\
& RN50$\times$16, ViT-L/14@336 & 49.84 & \textbf{58.57} & \underline{39.96} & \underline{55.28} & \textbf{63.92} & \textbf{45.35} & \underline{59.40} & \textbf{58.93}\\
& ViT-B/32,ViT-B/16 & 44.62 & 53.03 & 35.90 & 49.13 & 58.96 & 40.21 & 52.23 & 51.61\\
& ViT-B/32,ViT-L/14 & 47.19 & 56.27 & 38.14 & 52.75 & 62.07 &  42.69 & 56.66 & 55.54\\
& ViT-B/32,ViT-L/14@336px & 48.59 & 58.05 & 38.69 & 54.61 & \underline{63.45} & 43.28 & 57.80 & 57.48\\
& ViT-B/16,ViT-L/14 & 48.18 & 57.49 & 39.33 & 53.66 & 62.38 & 43.36 & 57.56 & 57.45 \\
& ViT-B/16,ViT-L/14@336px & \underline{49.86} & \underline{58.41} & 39.92 & \textbf{55.35} & 62.43 & 44.34 & 59.05 & \underline{58.82} \\
& ViT-L/14,ViT-L/14@336px & 48.82 & 57.03 & \textbf{40.35} & 53.62 & 60.65 & 44.04 & 59.03 & 58.27\\

\bottomrule \\
\end{tabular}%
\end{center}
\caption{Backbone ablation on \textbf{Referring Expressions Detection.} We compare CLIP backbones and their ensembles for ReCLIP~\cite{subramanian2022reclip} (without using relations resolution) and our Red Circle. The best and  second best for each method are \textbf{bolded} and \underline{underlined}, respectively.}%
\label{tab:ref-exp-all-backbones}	
\end{table*}
In this supplementary material, we provide more details about the datasets we use, implementation details and ablations, as well as further qualitative and quantitative evaluations.

\section{Datasets}

As noted in the main paper, we contribute additional annotations to the Spair71k dataset for some of our experiments.
We start from their keypoint annotations, which have no keypoint name annotations in the original dataset.
We then manually name all keypoints of the animal classes in Spair71k, as shown in~\cref{tab:animal_parts}.
We purposefully leave out some point annotations:
\begin{itemize}
\itemsep0em
    \item All animals have a left and right nostril annotated --- we take the right one in all classes and annotate it as \emph{nose}, and leave the left nostril out.
    \item All tails have point annotations at the start of the tail (attached to the body) and end of the tail. Because of the lack of words to precisely describe both points, we take the point \emph{not} attached to the body and annotate it as \emph{tail}, and leave the other one out.
    \item All ears have point annotations at the start of the ear (attached to the head), and at the pointy end. Because of the lack of words to precisely describe both points, we take the point \emph{not} attached to the head and annotate it as \emph{ear}, and leave the other one out.
    \item Birds have annotations for (i) foot, (ii) ankle, (iii) knee, which are often ambiguous and very close together. We only keep the \emph{foot} annotation.
\end{itemize}

Note we explicitly define different names for keypoints that can be ambiguous, e.g. eyes, ears, legs, etc. This ensures the role of questions and answers in~\cref{s:tasks} is satisfied.

\section{Discovered annotations}
Out of the discovered annotations in YFCC-15M, 44\% contain red circles. Overall, 73\% of the annotations were circles, and the rest were rectangles. 65\% of all annotations were red, 10\% yellow, 7\% blue, 7\% white, and the rest were  black, green, and purple.

\section{Additional implementation details}

\subsection{Referring Expressions Detection.}
\paragraph{Backbone}
We base the evaluation of our method on ReCLIP~\cite{subramanian2022reclip}, where an ensemble of two CLIP backbones is used --- RN50x16 and ViT-B/32.
We evaluated ReCLIP for all combinations of CLIP backbones in~\cref{tab:ref-exp-all-backbones} and found that, on average, this is the highest-performing one.
Similarly, for our method, we choose the ensemble of two backbones that lead to the highest performance --- RN50x16 and ViT-L/14@336.
Full comparison between the backbones can be found in~\cref{tab:ref-exp-all-backbones}.
\paragraph{Annotations}
We experiment with different marker shapes, sizes, and colours, and present the results in~\cref{tab:ablate_ref_det_marker}.
We find that, on average, a thin red circle leads to the best performance.
We use an ensemble of the red circle annotation and two additional augmentations --- blurring and gray-scaling the outside of the circle, for a total of three images per annotation, as shown.
These augmentations were inspired by examples in YFCC15M we discovered that were annotated like that.
We found that adding augmentations improves overall results.
However, we did not explore including augmentations beyond these.
We ablate these choices in~\cref{tab:ablate_ref_det_subtract_ensemble}.

\paragraph{Additional details}

We augment the text queries by prepending \textit{``This is''}. When subtracting the average with respect to other referring expressions, we use $Q=500$ randomly sampled expressions.

\subsection{Keypoint tasks}

\paragraph{Backbone}
We evaluate different backbones in Table 3 in the main paper
and find that ViT-L/14@336 performs best.

\paragraph{Annotations} We show examples of the markers we use in Fig.~4 in the main paper
. We compare a large range of sizes and colors, as shown in Table 2 in the main paper.
We find that a circle is the best marker, and drawing a cross over the point of interest is the worst.
The best-performing marker out of all is a red circle, which is the one we end up using.
In~\cref{fig:ablate-sizes-colors} we show a more detailed comparison of different colors, diameters, and thicknesses when using a circle annotation.
We see that a thin red circle is the best-performing marker.
We show what that circle looks like on an image in~\cref{fig:localize-predictions1}.

Given this, we draw red circles over the images, with radius $r=0.06H$ and thickness $t=0.01H$, where H is the shorter side of the image.
For the backbone we use, where the input size has $H=336$px, this becomes $r=20$px and $t=3$px.
\paragraph{Additional details} For the keypoint localization task, we set $M=30$, for a total of $30 \times 30 = 900$ query locations before applying the pseudo mask. The templates we use are \textit{``This is the \{part\} of a bird''} for CUB and \textit{``This image shows the \{part\} of the \{animal\}''} for SPair71k. We use a temperature parameter $\tau=\frac{1}{150}$.

\section{Qualitative evaluations}

We present qualitative evaluations on naming keypoints in~\cref{fig:naming-keypoints-1,fig:naming-keypoints-2}, keypoint localization in~\cref{fig:localize-predictions1,fig:localize-predictions2} and referring expressions comprehension in~\cref{fig:rec-predictions}.

\begin{table*}[ht]
\centering
\resizebox{0.95\textwidth}{!}{%
\begin{tabular}{c|llllll}
\toprule
\textbf{Part No} & \textbf{Bird} & \textbf{Cat} & \textbf{Cow} & \textbf{Dog} & \textbf{Horse} & \textbf{Sheep}  \\
\midrule
0 & crown & --- & --- & --- & --- & --- \\
1 & right wing & --- & --- & --- & --- & ---  \\
2 & left wing & right ear & right ear & right ear & right ear & right ear  \\
3 & beak & left ear & left ear & left ear & left ear & left ear  \\
4 & --- & right eye & right eye & right eye & right eye & right eye \\ 
5 & --- & left eye & left eye & left eye & left eye & left eye \\ 
6 & forehead & nose & nose & nose & nose & nose  \\
7 & right eye & --- & --- & forehead & --- & ---  \\
8 & left eye & mouth & mouth & mouth & mouth & mouth  \\
9 & nape & front right paw & front right hoof & front right paw & forehead & front right hoof  \\
10 & right foot & front left paw & front left hoof & front left paw & front right hoof & front left hoof  \\
11 & left foot & hind right paw & hind right hoof & hind right paw & front left hoof & hind left hoof  \\
12 & --- & hind left paw & hind left hoof & hind left paw & hind right hoof & hind right hoof  \\
13 & tail & tail & tail & tail & hind left hoof & tail  \\
14 & --- & --- & --- & --- & tail & ---  \\
15 & --- & --- & front right knee & neck & --- & front right knee  \\
16 & --- & --- & front left knee & --- & front right knee & front left knee  \\
17 & --- & --- & hind right knee & --- & front left knee & hind right knee  \\
18 & --- & --- & hind left knee & --- & hind right knee & hind left knee  \\
19 & --- & --- & right horn & --- & hind left knee & right horn  \\
20 & --- & --- & left horn & --- & --- & ---  \\
\bottomrule \\
\end{tabular}}
\caption{\textbf{Part names for keypoint annotations of the SPair71k dataset}. Part No is the part number in the SPair71k annotations. Some parts are annotated inconsistently in the original annotations, e.g. ``tail'' is part number 10 for the ``horse'' class, but part number 9 for all other animal classes.}%
\label{tab:animal_parts}
\end{table*}

\begin{figure*}[]
\centering
\includegraphics[width=0.9\textwidth]{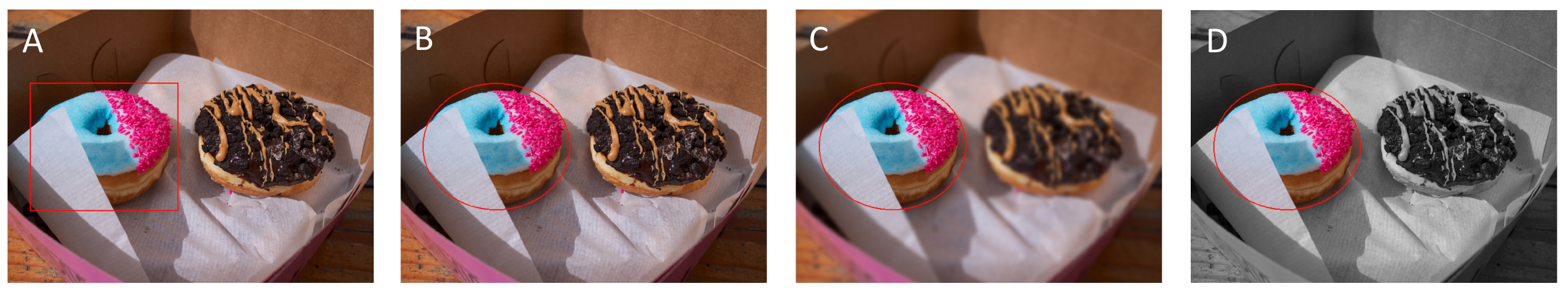}
\caption{\textbf{Annotations for Referring Expressions Detection.} Here we show the annotation types we consider.
A\@: original bounding box annotation.
B\@: Red Circle.
C\@: Red Circle + Blur outside.
D\@: Red Circle + Gray outside. In our experiments, we use an ensemble of B, C and D unless stated otherwise.}%
\label{fig:app-anno-rec}
\end{figure*}

\begin{table*}
\begin{center}
\begin{tabular}{cccccc|ccc|cc}
\toprule
\multicolumn{3}{c|}{Component}  &\multicolumn{3}{c|}{\textbf{RefCOCO}} & \multicolumn{3}{c|}{\textbf{RefCOCO+}} & \multicolumn{2}{c}{\textbf{RefCOCOg}} \\
Red Circle & Subtract & Ensemble &  Val & TestA & TestB & Val & TestA & TestB & Val & Test \\
\midrule
\cmark & \xmark & \xmark & 42.01 & 48.58 & 36.90 & 47.55 & 53.56 & 41.05& 50.84 & 51.47 \\
\cmark & \cmark & \xmark & 43.67 & 50.20 & 38.59 & 48.98 & 54.70 & 43.06 & 54.29 & 52.98 \\
\cmark & \cmark & \cmark & \textbf{49.84} & \textbf{58.57} & \textbf{39.96} & \textbf{55.28} & \textbf{63.92} & \textbf{45.35} & \textbf{59.40} & \textbf{58.93} \\
\bottomrule \\
\end{tabular}%
\end{center}
\caption{\textbf{Ablation study.} We ablate subtracting the mean wrt negative queries and ensembling different marker types (red circle + red circle and blur outside + red circle and grey outside).
Here we use RN50$\times$16 and ViT-L/14@336px backbones and a red circle with the optimal size described in~\cref{tab:ablate_ref_det_marker}}%
\label{tab:ablate_ref_det_subtract_ensemble}
\end{table*}

\begin{table*}
\begin{center}
\begin{tabular}{lllccc|ccc|cc}
    \toprule
\multicolumn{3}{c|}{Annotation Type}  &\multicolumn{3}{c|}{\textbf{RefCOCO}} & \multicolumn{3}{c|}{\textbf{RefCOCO+}} & \multicolumn{2}{c}{\textbf{RefCOCOg}} \\
 Shape & Color & Size &  Val & TestA & TestB & Val & TestA & TestB & Val & Test \\
    \midrule
Circle & Red & 1 & 38.7 & \underline{45.1} & \underline{34.0} & \underline{44.4} & \textbf{50.0} & \underline{39.1} & 48.1 & \textbf{50.0} \\
Circle & Red & 2 & 32.2 & 35.9 & 29.1 & 37.6 & 40.9 & 33.5 &  45.3 & 46.4 \\
Circle & Red & 4 & 37.4 & 43.6 & 31.5 & 43.3 & 47.8 & 37.3 & 43.7 & 48.0 \\
Circle & Red & 8 & 36.3 & 42.6 & 31.3 & 42.1 & 47.3 & 36.3 & 45.2 & 45.4 \\
Rectangle  & Red & 1 & 35.1 & 38.3 & 33.5 & 39.2 & 41.4 & 37.3 & 44.3 & 43.4 \\
Rectangle & Red & 2 & 35.1 & 38.3 & 33.2 & 39.1 & 41.8 & 37.3 & 44.8 & 44.1 \\
Rectangle & Red & 4 & 34.1 & 37.8 & 32.3 & 39.0 & 41.3 & 36.5 & 43.7 & 44.1 \\
Rectangle & Red & 8 & 33.7 & 37.6 & 32.7 & 37.9 & 40.3 & 34.9 & 41.1 & 40.1 \\
Circle & Green & 1 & \textbf{39.3} & \textbf{45.4} & \textbf{34.8} & 43.8 & \underline{49.9} & 38.1 & 47.2 & 47.4 \\
Circle & Purple & 1 & \underline{38.9} & 44.8 & 34.0 & \textbf{44.5} & 49.4 & \textbf{39.2} & \textbf{49.5} & \underline{49.2} \\
Circle & Blue & 1 & 37.7 & 44.9 & 33.5 & 43.4 & 49.1 & 37.3 & 48.2 & 48.3 \\
Circle & Yellow & 1 & 38.5 & 44.1 & 34.6 & 43.7 & 49.0 & 38.9 & \underline{48.6} & 48.1 \\
    \bottomrule \\
\end{tabular}%
\end{center}
\caption{\textbf{Comparison of different sizes, shapes, colors.} A unit size of 1 corresponds to 0.5\% of the larger side of the image, which is 1 pixel for an image of size 224. Here we do \emph{not} use ensembling and subtraction of the mean wrt other queries in order to evaluate the effectiveness of different markers themselves. The best and  second best are \textbf{bolded} and \underline{underlined}, respectively.}
\label{tab:ablate_ref_det_marker}
\end{table*}

\begin{figure*}[t]
\centering
\includegraphics[width=0.95\textwidth]{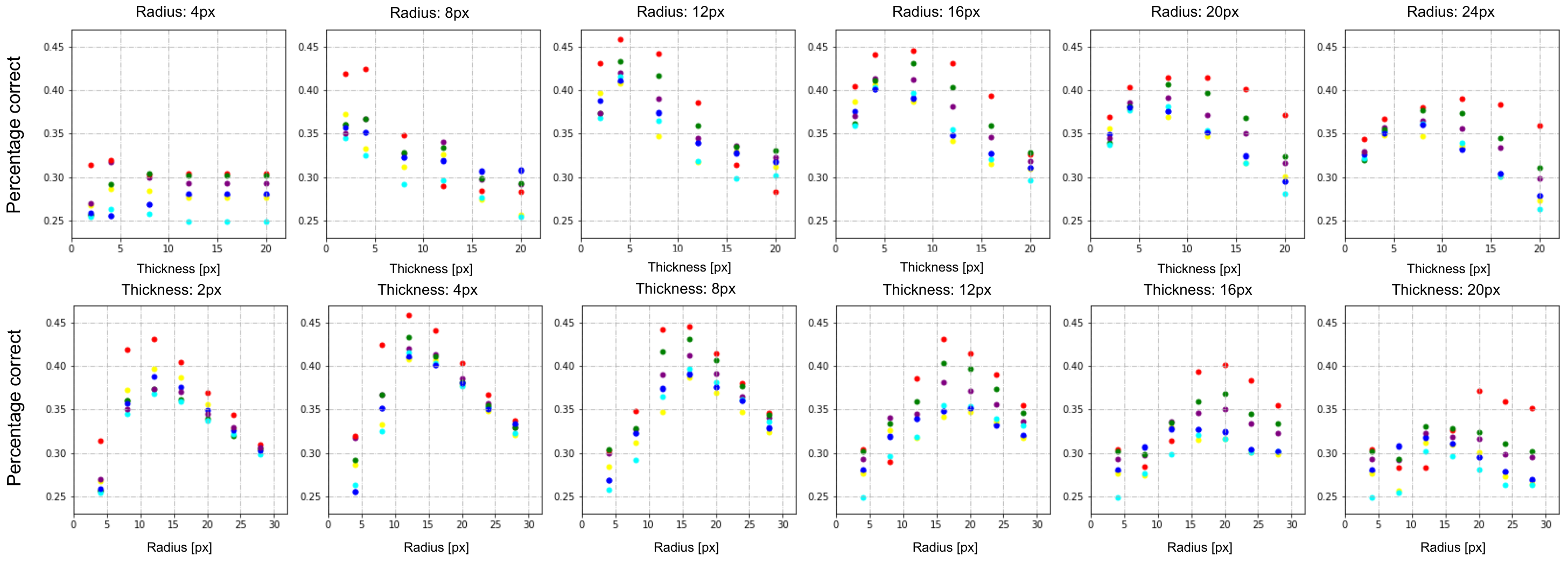}
\caption{\textbf{Ablation of circle sizes and colours for keypoint matching.} We present results on the CUB dataset when varying the diameter and thickness of the annotations. The presented numbers are for text-to-image matching. The best performing annotation has a radius of 12px and thickness of 4px. The colour of the dots on the scatter plot illustrates the colour of the annotation --- red, green, blue purple, yellow, cyan.}%
\label{fig:ablate-sizes-colors}
\end{figure*}

\begin{figure*}[t]
\centering
\includegraphics[width=0.8\textwidth]{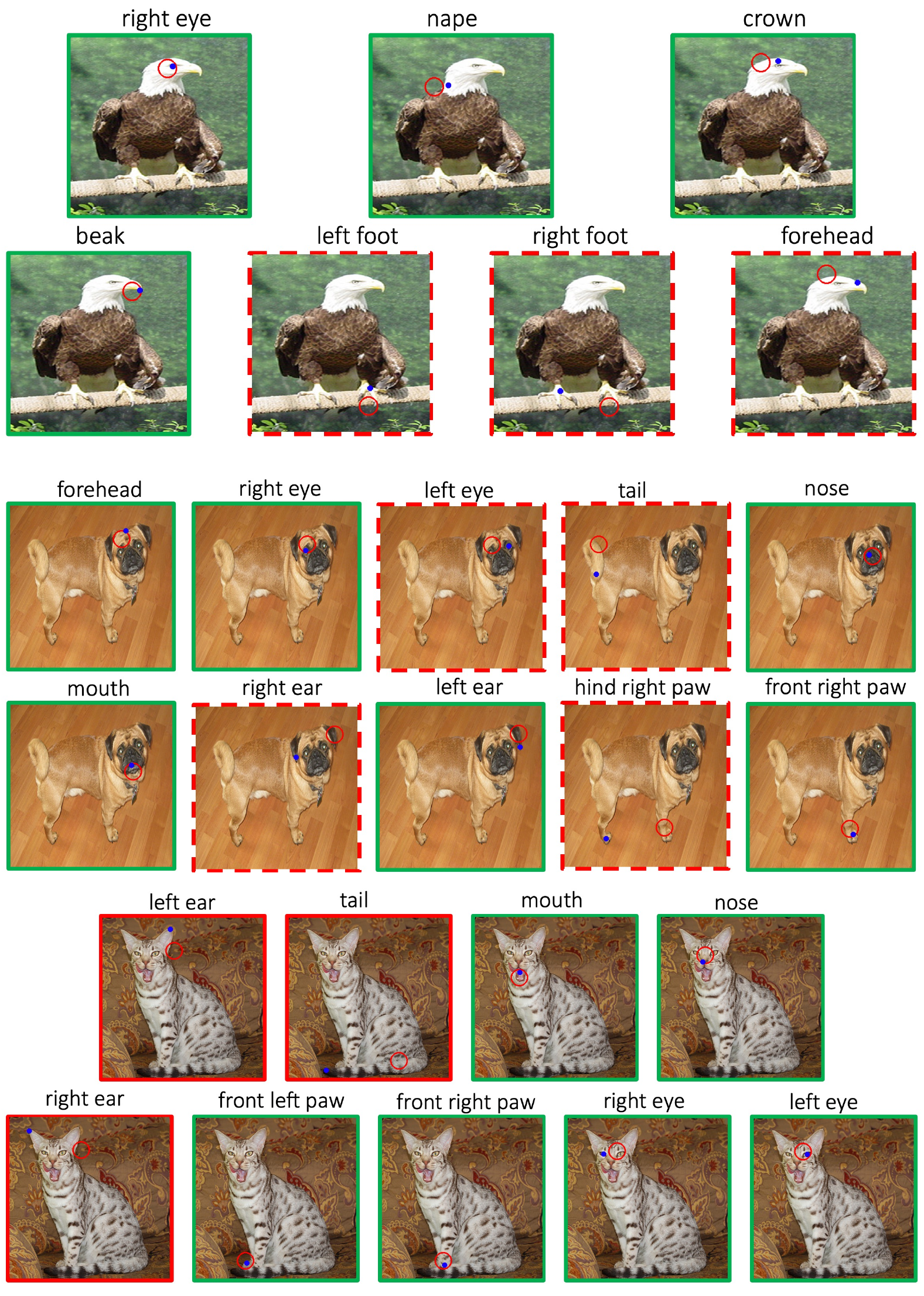}
\caption{\textbf{Qualitative evaluation of keypoint localization on SPair71k.} We show all keypoint names for the images and color code in green and red (dashed) the correct and wrong localizations according to PCK with $\alpha=0.1$. The red circle is the marker we use and the blue dot is the ground truth location.}%
\label{fig:localize-predictions1}
\end{figure*}

\begin{figure*}[t]
\centering
\includegraphics[width=0.8\textwidth]{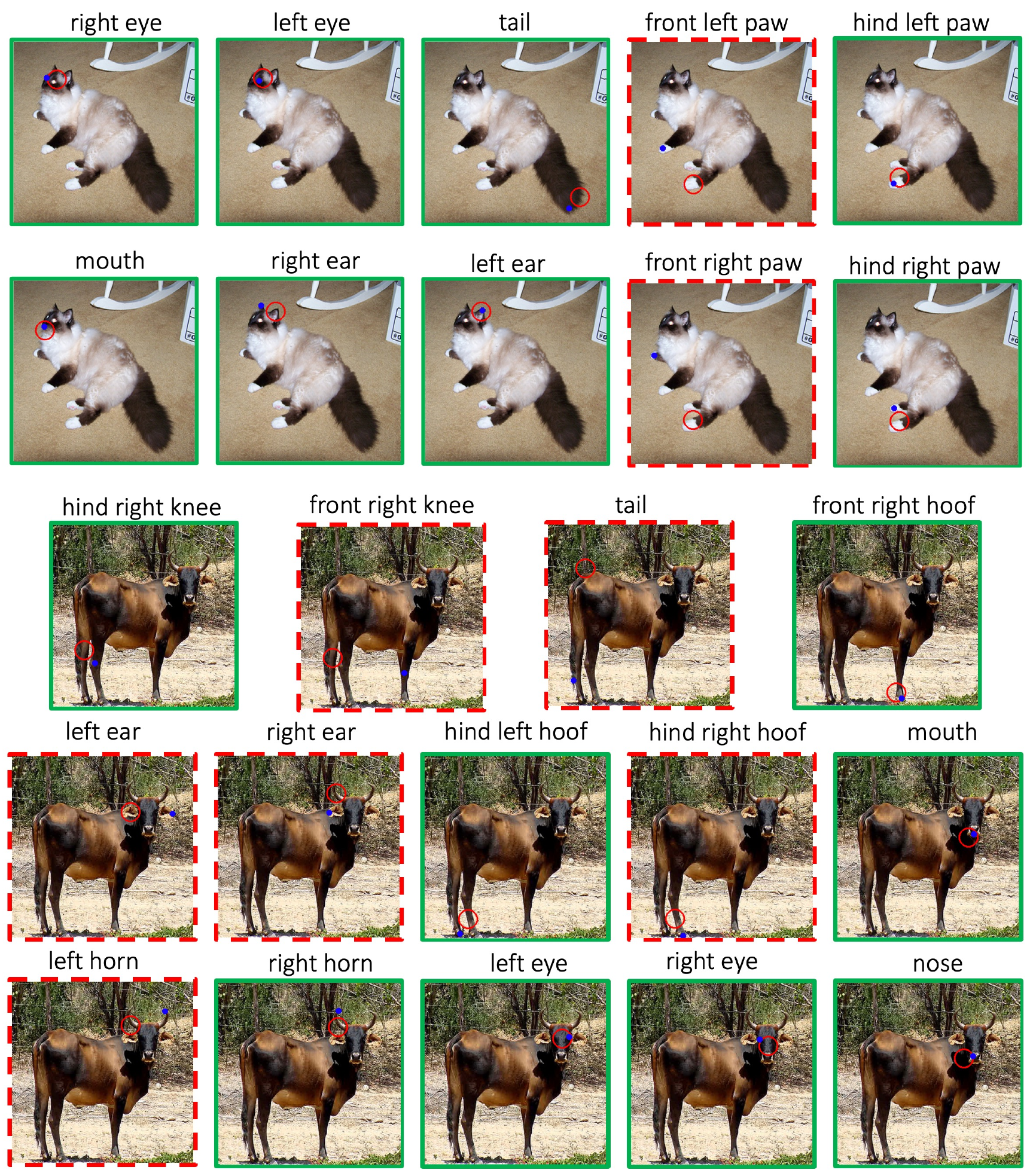}
\caption{\textbf{Qualitative evaluation of keypoint localization on SPair71k.} We show all keypoint names for the images and color code in green and red (dashed) the correct and wrong localizations according to PCK with $\alpha=0.1$. The red circle is the marker we use and the blue dot is the ground truth location.}%
\label{fig:localize-predictions2}
\end{figure*}

\begin{figure*}[t]
\centering
\includegraphics[width=0.95\textwidth]{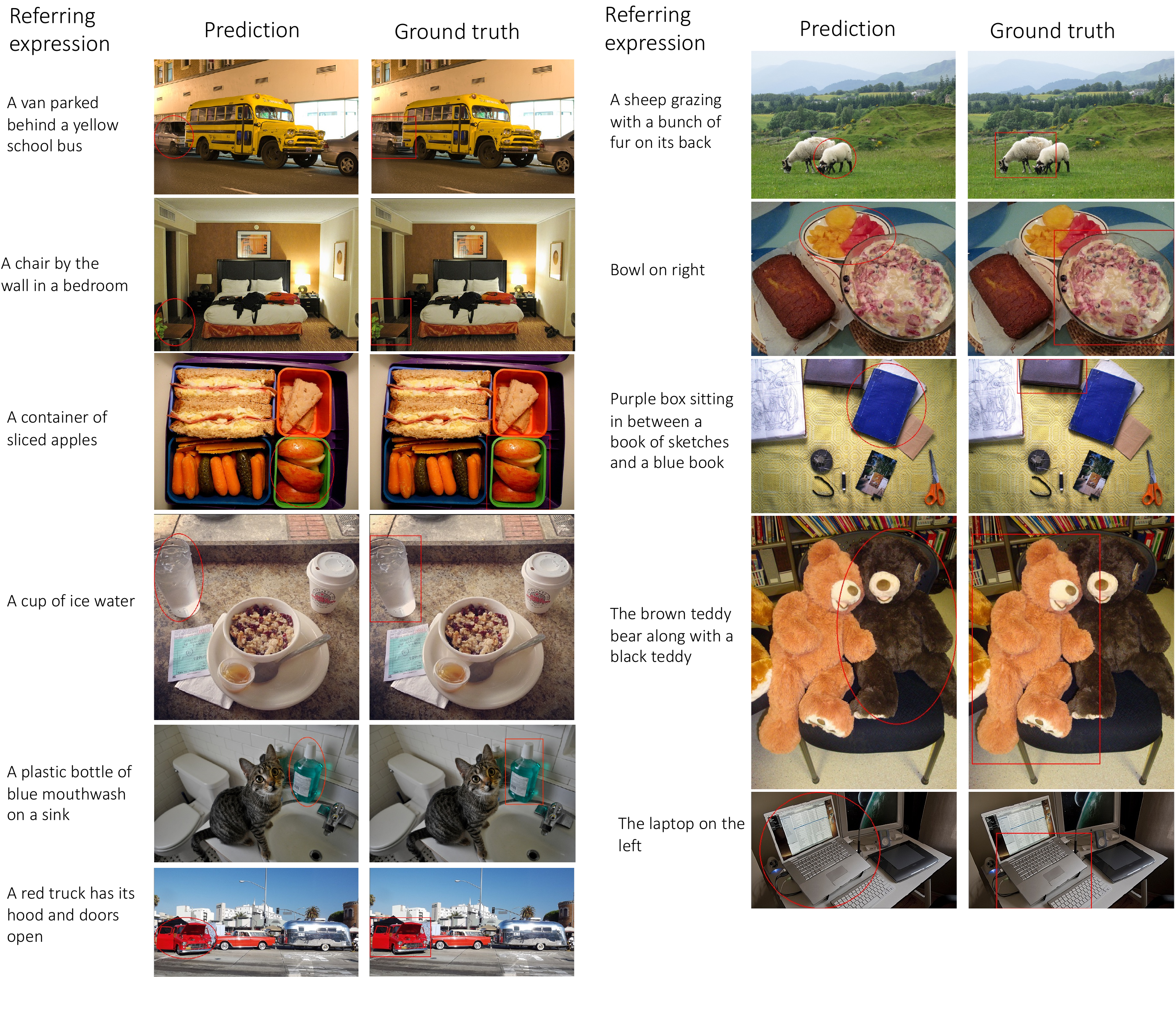}
\caption{\textbf{Qualitative results on REC on the RefCOCOg dataset.} Left: correct predictions. Right: wrong predictions. The last row on the right shows an example where the ground-truth bounding box is wrong.}%
\label{fig:rec-predictions}
\end{figure*}

\begin{figure*}[t]
\centering
\includegraphics[width=0.95\textwidth]{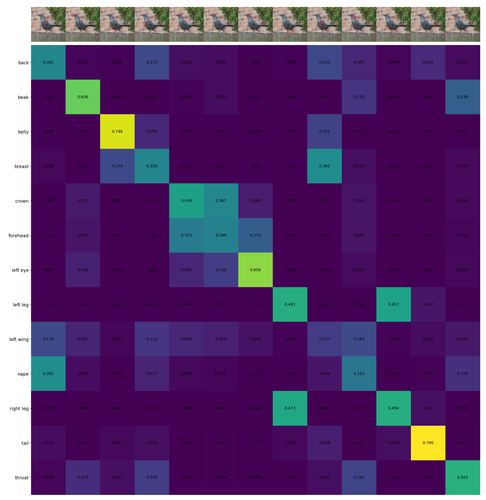}
\caption{\textbf{Naming keypoints.}Normalized cost matrix for an image from CUB}%
\label{fig:naming-keypoints-1}
\end{figure*}

\begin{figure*}[t]
\centering
\includegraphics[width=0.95\textwidth]{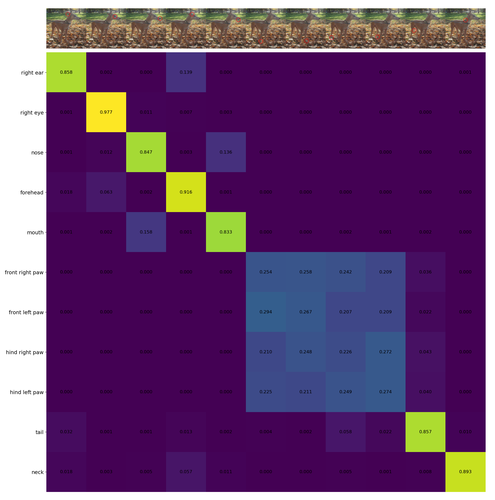}
\caption{\textbf{Naming keypoints.} Normalized cost matrix for an image from SPair71k}%
\label{fig:naming-keypoints-2}
\end{figure*}

\end{document}